\documentclass[journal]{IEEEtai}

\usepackage[colorlinks,urlcolor=blue,linkcolor=blue,citecolor=blue]{hyperref}

\usepackage{color,array}

\usepackage{times}
\usepackage{epsfig}
\usepackage{amsmath}
\usepackage{amssymb}
\usepackage{booktabs}
\usepackage{subfigure}

\usepackage{etoolbox}
\usepackage{mathrsfs}

\newcommand{\MyMapTemplatePrefix}[4]{\expandafter#1\csname#3#4\endcsname{#2{#4}}}
\newcommand{\MyMapTemplatePrefixNew}[5]{\expandafter#1\csname#4#5\endcsname{#2{#3{#5}}}}

\forcsvlist{\MyMapTemplatePrefix {\def} {\mathcal}{mc}} {A,B,C,D,E,F,G,H,I,J,K,L,M,N,O,P,Q,R,S,T,U,V,W,X,Y,Z}

\usepackage{graphicx}

\begin{document}

\title{Structure-Sensitive Graph Dictionary Embedding for Graph Classification}

\author{Guangbu Liu, Tong Zhang, Xudong Wang, Wenting Zhao, Chuanwei Zhou, and Zhen Cui, \IEEEmembership{Member, IEEE}
\thanks{Guangbu Liu, Tong Zhang, Xudong Wang, Wenting Zhao, Chuanwei Zhou, and Zhen Cui are with the Key Lab of Intelligent Perception and Systems for High-Dimensional, the School of Computer Science and Engineering, Nanjing University of Science and Technology, China (e-mail: \{liuguangbu, tong.zhang, xd\_wang, wtingzhao, cwzhou, zhen.cui\}@njust.edu.cn).}
}

\markboth{Journal of IEEE Transactions on Artificial Intelligence, Vol. 00, No. 0, Month 2020}
{First A. Author \MakeLowercase{\textit{et al.}}: Bare Demo of IEEEtai.cls for IEEE Journals of IEEE Transactions on Artificial Intelligence}

\maketitle

\begin{abstract}
Graph structure expression plays a vital role in distinguishing various graphs. In this work, we propose a Structure-Sensitive Graph Dictionary Embedding (SS-GDE) framework to transform input graphs into the embedding space of a graph dictionary for the graph classification task. Instead of a plain use of a base graph dictionary, we propose the variational graph dictionary adaptation (VGDA) to generate a personalized dictionary (named adapted graph dictionary) for catering to each input graph. In particular, for the adaptation, the Bernoulli sampling is introduced to adjust substructures of base graph keys according to each input, which increases the expression capacity of the base dictionary tremendously.  To make cross-graph measurement sensitive as well as stable, multi-sensitivity Wasserstein encoding is proposed to produce the embeddings by designing multi-scale attention on optimal transport. To optimize the framework, we introduce mutual information as the objective, which further deduces to variational inference of the adapted graph dictionary. We perform our SS-GDE on multiple datasets of graph classification, and the experimental results demonstrate the effectiveness and superiority over the state-of-the-art methods.
\end{abstract}

\begin{IEEEImpStatement}
Graphs have been widely used to model ubiquitous irregular data in various real-world and scientific fields. To learn robust representation from graphs, graph kernel and dictionary learning are flourishing in recent years. However, the fixed graph kernels and graph dictionaries in these methods are not flexible to handle abundant structure patterns and hence possess limited expression capacity. For this issue, we propose the variational graph dictionary adaptation method to improve graph representation ability. Specifically, our graph dictionary can be adaptive to various graph patterns and sensitive to cross-graph correlation, which provides a new framework for graph representation learning.
\end{IEEEImpStatement}


\begin{figure}[!thb]
	\centering
	\includegraphics[width=0.95\linewidth]{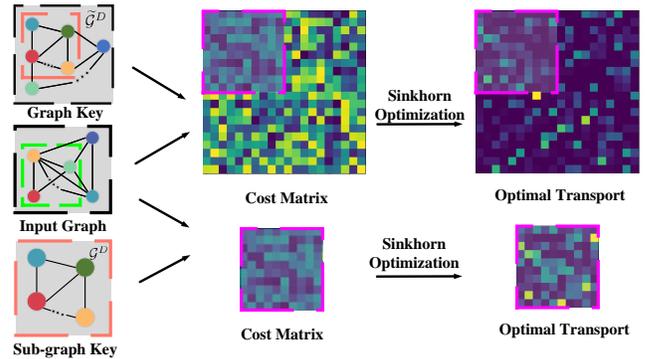}
	\caption{The visualization of optimal transport (OT) results between the same input graph and two graph keys. The orange-dotted boxes indicate the same structure in graph keys. The purple-dotted boxes in each column denote the corresponding cross-graph cost matrices and OT results between the input subgraph (green-dotted) and graph/subgraph key (orange-dotted), respectively.}
	\label{necessity_VGDA}
\end{figure}

\section{Introduction}

\IEEEPARstart{G}{raphs} are usually composed of one node set and one edge set, where nodes represent individual objects and edges denote relationships among them. Due to rather flexible structures, graphs have been widely used to model ubiquitous irregular data in various real-world and scientific fields, such as biological graphs~\cite{duvenaud2015convolutional,hamilton2017inductive} and social networks. To obtain powerful representation ability on graphs, graph deep learning is flourishing in recent years, and has made many milestone progresses on several graph-related tasks including community detection, document classification, et al. Among them, an essential issue is graph classification, which endeavors to learn the most discriminant representation of graphs in certain measurement metrics.

The existing graph classification methods generally fall into two main streams: graph kernel-based algorithms and graph neural networks (GNNs). As the traditional and classical representative, graph kernels~\cite{borgwardt2005shortest, shervashidze2009efficient}~measure similarities of graphs in a lifting high-dimensional space of structure statistics quantified by graphlets, where structural information is well preserved. As a contrast, GNNs~\cite{kipf2016semi, velivckovic2017graph, xu2018powerful} attempt to extract high-level and discriminative features by stacking multiple neural network layers, wherein the adjacent information is aggregated in an iterative manner. Hence, GNNs are capable of exploiting local structures of graph to some extent, and have achieved better results in graph classification due to the essence of deep architecture. 

Even though much considerable progress has been made by GNNs in graph classification, the over-smoothness~\cite{xu2018powerful} limits the expression ability of features because of the structure confusion during information aggregation. When retrospecting graph kernels again, the good preservation ability of graph structures is really fascinating to GNNs. Hereby, some deep graph dictionary learning methods~\cite{Zhang2021DeepWG, vincent2021online} encode input graph with graph dictionary in a GNN architecture, and meantime achieve promising performance. However, there are two major problems that remain to be addressed: i) a fixed graph dictionary usually has limited capacity to express giant structure patterns, as combinatorial-explosive graph structures with exponential-order magnitude would be overwhelming for such a dictionary; 
ii) the similarity measurement across input graph and graph dictionary key should be stable, also sensitive to local structural variations, which guarantees high representation ability on graph dictionary. 


To tackle the two issues above, in this work, we propose a Structure-Sensitive Graph Dictionary Embedding (SS-GDE) framework to facilitate graph representation modeling for graph classification. To fully liberate the graph dictionary, we propose variational graph dictionary adaptation (VGDA) to conduct individual structure selections from graph dictionary keys for each input graph. Such a selective adjustion tremendously expands the capacity of the original fixed graph dictionary (called base graph dictionary, BGD), and generates a personalized specific dictionary adapted for each input graph, which we call adapted graph dictionary (AGD) versus base graph dictionary. To effectively choose the corresponding substructures from base graph keys, we introduce a Bernoulli sampling to be learnt during the variational inference in VGDA. For the measurement between the input graph and adaptive dictionary keys, we employ the cross-graph Wasserstein distance that calculates optimal transport (OT) between graphs, which is rather stable as verified in ~\cite{Zhang2021DeepWG}. Hence, compared with the fixed graph dictionary, the VGDA is advantageous in better exploring significant structural information through generating adaptive substructures (the representation difference between one graph key and its substructure is shown in Fig.~\ref{necessity_VGDA}) with better expression ability for each input. 
But to increase the sensitivity for cross-graph correlation, we propose multi-sensitivity Wasserstein encoding (MS-WE) by introducing multi-scale attention on OT, which could adaptively capture those important local correlation patterns for the final accurate representation of input graph. To optimize the proposed framework, we introduce mutual information as the objective, which further deduces to variational inference of adapted graph dictionary. To evaluate the SS-GDE framework, extensive experiments are conducted on multiple graph classification datasets, and the experimental results validate the effectiveness and the superiority over the state-of-the-art methods. 

In summary, the contributions of our work are four-fold: i) propose a structure-sensitive graph dictionary embedding framework to promote deep graph learning for graph classification; ii) propose variational graph dictionary adaptation to release the potential capacity of base graph dictionary; iii) design multi-sensitivity Wasserstein encoding to guarantee the sensitivity as well as stability of cross-graph measurement; iv) report new state-of-the-art results on some datasets.

\section{Related work}

In this section, we first review the previous methods of graph classification, then introduce works related to inherently interpretable models and Wasserstein distance learning.

\textbf{Graph Classification.} Many recent techniques have been proposed to solve the graph classification problem. Some early approaches dedicated to building kernel functions to measure similarities among graphs. These kernel-based methods decompose graphs into sub-structure such as random-walks~\cite{gartner2003graph}, shortest path~\cite{borgwardt2005shortest}, graphlets~\cite{shervashidze2009efficient} and subtrees~\cite{shervashidze2011weisfeiler}.
GNN can directly operate on graph-structured data to extract expressive graph-level representation by stacking multiple neural network layers, which can aggregate neighbor node features and have achieved promising performance in the graph classification task. Besides, various convolution~\cite{kipf2016semi, luo2017deep, niepert2016learning} and pooling operations were proposed to learn robust node features and graph representations in recent years. Graph Convolutional Network(GCN)~\cite{kipf2016semi} proposed a layer-wise propagation rule based on a first-order approximation of spectral convolution on graphs via the Chebyshev polynomial iteration. Graph Attention Network (GAT)~\cite{velivckovic2017graph} highlighted more information nodes by assigning different weights to different nodes in the neighborhood. In addition to the above-mentioned methods, many pooling strategies have emerged, which can be categorized as node selection~\cite{lee2019self, li2020graph, nouranizadeh2021maximum}, graph coarsening~\cite{ying2018hierarchical, yuan2020structpool} and other methods~\cite{gmt_iclr21,li2019semi}. 

\begin{figure*}[!thb]
	\begin{center}
		\includegraphics[width=1\linewidth]{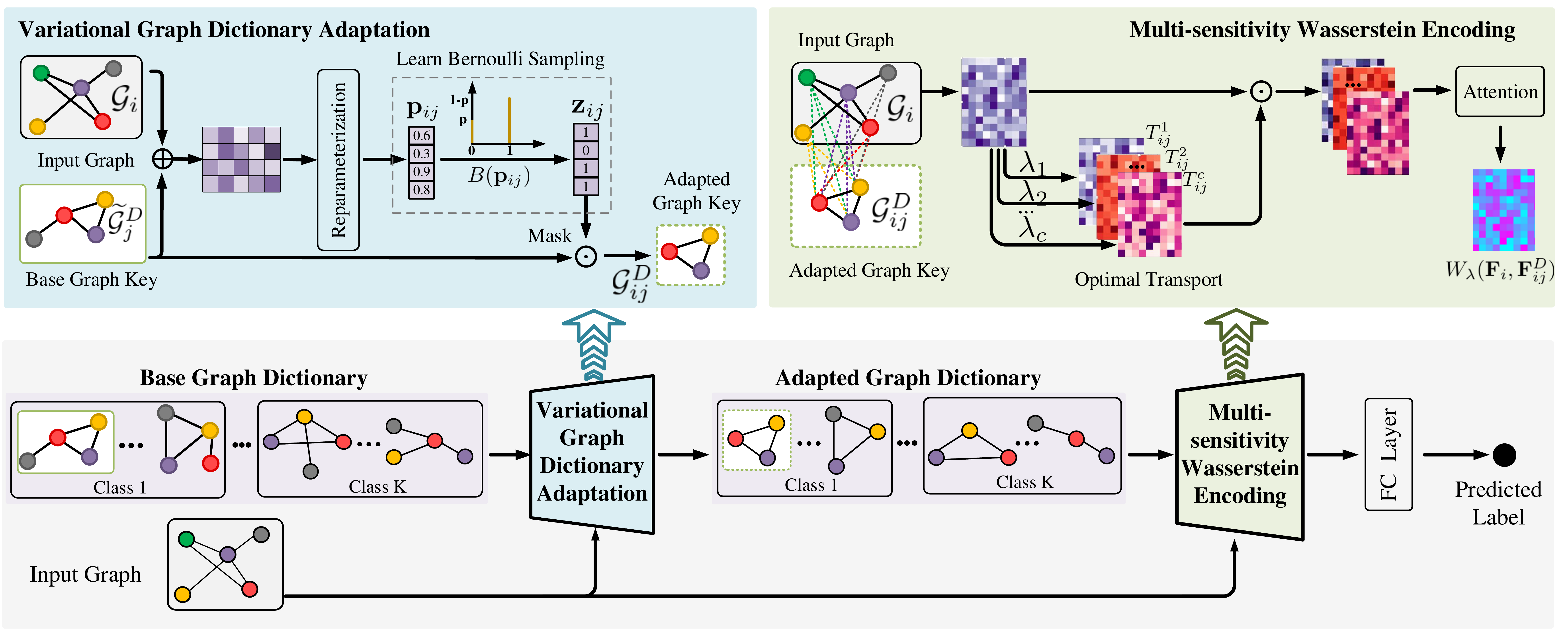}
	\end{center}
	\caption{The architecture of the proposed SS-GDE framework. It contains two main modules named the variational graph dictionary adaptation (VGDA) and multi-sensitivity Wasserstein encoding (MS-WE). Given one input graph, as the first step, a base graph dictionary (BGD) is constructed to support the subsequent embedding. Then, the input graph and BGD are fed into the VGDA module to learn the adapted graph dictionary (AGD) corresponding to each input. In this process, the Bernoulli sampling is learned to extract substructures from the BGD by cross-correlating input and its dictionary keys. Next, the learned AGD is fed into the MS-WE module to produce the embedding of the input graph. To make better cross-graph embedding, the MS-WE employs the multi-sensitivity regularization before the OT to improve the sensitivity of structure variations, and further introduces the attention mechanism to capture salient structural patterns. Finally, the obtained embeddings pass through fully-connected layers for low-dimensional representations to facilitate the classification. More details can be found in the main body.}
	\label{framework}
\end{figure*}

\textbf{Wasserstein Distance Learning.} Wasserstein distance measures the difference between two probability distributions defined on a given metric space by leveraging the OT principle and has been widely used in machine learning and pattern matching fields. 
Numerous algorithms~\cite{becigneul2020optimal, chen2020graph, frogner2015learning, titouan2019optimal, vincent2022template} were proposed to learn representation from graphs or measure this distance. For instance, ~\cite{togninalli2019wasserstein} proposed that calculate the Wasserstein distance between the node feature vector distributions of two graphs to find subtler differences. Fused Gromov-Wasserstein (FGW)~\cite{titouan2019optimal} was introduced to consider both features and structure information in the OT problem. Some other works~\cite{rolet2016fast, schmitz2018wasserstein, vincent2021online} attempted to conduct dictionary learning using W-distance or GW-distance for fitting data term. \cite{vincent2021online} proposed a linear dictionary learning approach to model graphs by computing inter-graph correlation through GW-distance.

\textbf{Inherently Interpretable Models.} The interpretability of models refers to understanding their internal mechanism. Many works~\cite{dai2021towards, lin2022orphicx, luo2020parameterized, miao2022interpretable, wu2022discovering, ying2019gnnexplainer, yu2022improving, yu2020graph} have been proposed to extract meaningful data patterns for prediction by post-hoc methods~\cite{luo2020parameterized, ying2019gnnexplainer} as well as inherently interpretable models~\cite{dai2021towards, miao2022interpretable, wu2022discovering}. However, To make the model more interpretable, it may need to sacrifice prediction performance. GSAT~\cite{miao2022interpretable} introduced stochastic attention to block the task-irrelevant graph components information while learning stochasticity-reduced attention to select task-relevant subgraphs to provide the interpretability of the model. DIR~\cite{wu2022discovering} proposed to create multiple distributions by conducting interventions to the training distribution and to filter out the spurious and unstable patterns. 

Compared to those existing methods above, our SS-GDE has apparently different aspects:
$(1)$ Instead of a plain use of base graph dictionary, we introduce the Bernoulli sampling to adjust substructures of base graph keys to generate a personalized dictionary adapted for each input graph. Such a selective adjustion significantly expands the capacity of original fixed graph dictionary. Furthermore, we introduce mutual information as objective, which deduces to the variational inference of adapted graph dictionary.
$(2)$ We design the MS-WE module to use cross-graph Wasserstein distance for the stability of embedding and further introduce multi-sensitivity regularization to improve the sensitivity of structure variations. Hence, those important local correlation patterns could be well captured for the accurate representation of input graphs.

\section{The proposed Method}
In the following parts, vectors/matrices are denoted with lowercase/uppercase letters in boldface, and $\Box^\intercal$ represents the transpose. A calligraphic symbol may either indicate a tuple, e.g. the graph tuple $\mathcal{G}_i$ consisting of node/edge sets, or simply a set, e.g. a node set $\mathcal{V}_i$. $\Box^D$ means that the matrix/vector corresponds to the graph dictionary, e.g. the base graph dictionary is denoted as $\widetilde{\mathcal{D}}=\{\widetilde{\mathcal{G}}^D_1,\cdots,\widetilde{\mathcal{G}}^D_K\}$  and the adapted graph dictionary is denoted as $\mathcal{D}=\{\mathcal{G}^D_1,\cdots,\mathcal{G}^D_K\}$ (K is the number of the graph keys).
In this section, we first overview the whole architecture of our proposed SS-GDE framework, then describe those main learning processes in detail.

\subsection{Overview}
The whole architecture of the proposed SS-GDE framework is shown in Fig.~\ref{framework}. In general, it contains two main learning modules: variational graph dictionary adaptation (VGDA)  and multi-sensitivity Wasserstein encoding (MS-WE). Before VGDA, a base graph dictionary (BGD) $\widetilde{\mathcal{D}}$ is first constructed as detailed in Section ~\ref{sec:EXsetup}
to support the subsequent embedding representation of the input graph. Specifically, it is optimizable as model parameters and initialized using the training samples following the same process in~\cite{Zhang2021DeepWG}. For the rough graphs in the dictionary or input set, graph convolution neural networks (GCNNs) may be used to learn the primary expression of each node.
Given an input graph $\mathcal{G}$, the VGDA learns an adapted graph dictionary $\mathcal{D}$ from the base dictionary through learning Bernoulli sampling during cross-correlating input and dictionary keys.
Such a process generates more expressive structural dictionaries for the next embedding representation. The detail of VGDA could be found in Section~\ref{sec:VGDA}. The adapted graph dictionary $\mathcal{D}$ is fed into the MS-WE module to produce the embedding of input graph. To make better cross-graph embedding, in MS-WE, we use cross-graph Wasserstein distance for the stability of embedding and introduce multi-sensitivity regularization for the sensitivity of structure variations. The detail of VGDA could be found in Section~\ref{sec:MS-WE}. Finally, the resulting embeddings pass through fully-connected layers for low-dimensional representations. To optimize the SS-GDE framework, mutual information is introduced as the objective, which deduces the variational inference of the AGD, as derived in Section~\ref{sec:MI}. In the training process, the whole architecture as well as the base graph dictionary can be optimized in an end-to-end tuning mode through back-propagation.

\subsection{The Objective Function}\label{sec:MI}
We first describe the objective function of the proposed SS-GDE for a clear explanation of the whole architecture. Generally, the SS-GDE aims for a supervised graph classification task. For the formulation, the whole embedding process is denoted as $f_{\Psi,\phi}:(\mathcal{G},\widetilde{\mathcal{D}})\rightarrow \mathcal{Y}$. It means to project input samples in graph space $\mathcal{G}$ into the label space $\mathcal{Y}$ based on the BGD $\widetilde{\mathcal{D}}$. Here, we employ mutual information denoted as $I(\cdot)$ to measure the relationship between the learned embeddings and their corresponding labels. Hence, the objective function to maximize is defined as:
\begin{align} 
	\label{obj} L = I(\mathbf{y}, f_{\Psi,\phi}(\mathcal{G}, \widetilde{\mathcal{D}})).
\end{align}
As mentioned above, an AGD $\mathcal{D}$ is further learned from $\widetilde{\mathcal{D}}$ through Bernoulli sampling with the factor set denoted as $\mathcal{Z}=\{\mathbf{z}_{i1}, \cdots, \mathbf{z}_{iK}\}$. Hence, we divide all those parameters of our framework into two sets, including (i) one set denoted as $\phi$, which is specifically used to learn the Bernoulli factor set $\mathcal{Z}$, and (ii) the set of the other parameters, denoted as $\Psi$. Then, the optimization objective in Eqn.~(\ref{obj}) could be further bounded:
\begin{equation}
	\begin{aligned} 
		\label{MI_1}  &I(\mathbf{y}, f_{\Psi,\phi}(\mathcal{G}, \widetilde{\mathcal{D}}))
		\ge E_\mathbf{y|\mathcal{G},\widetilde{\mathcal{D}}}( E_{q_{\phi}(\mathcal{Z})} (\log p_{\Psi}(\mathbf{y}|\mathcal{G}, \mathcal{D})) \\
		&\qquad\qquad\qquad\qquad\quad- D_{KL}(q_{\phi}(\mathcal{Z})||p_{\phi}(\mathcal{Z}|\mathcal{G}, \widetilde{\mathcal{D}})).
	\end{aligned}
\end{equation}
The detailed derivation can be found in Appendix. 

$p_{\Psi}(\mathbf{y}|\mathcal{G}, \mathcal{D})$ means to infer the predicted label based on each input sample and the corresponding AGD. It mainly involves the MS-WE (in Section~\ref{sec:MS-WE}) and fully-connected layers. 
$q_\phi(\mathcal{Z})$ is the expected Bernoulli distribution, while $p_{\phi}(\mathcal{Z}|\mathcal{G}, \widetilde{\mathcal{D}})$ is the conditional probability of the factor set $\mathcal{Z}$ which is usually intractable. Specifically, for each input $\mathcal{G}_i$, we assume that those sampling factors $\mathbf{z}_{i1}, \cdots, \mathbf{z}_{iK}$ are independent and identically distributed, and each of them conforms the Bernoulli distribution. Hence, the KL divergence can be rewritten as:
\begin{align}\label{KL}
	\nonumber D_{KL}(q_{\phi}(\mathcal{Z})||p_{\phi}(\mathcal{Z}|\mathcal{G}, \widetilde{\mathcal{D}}))~~~~~~~~~~~~~~~~~~~~~~~~~~~~~\\
	= \sum_{i} \sum_{j=1}^{K} D_{KL}(q_{\phi}(\mathbf{z}_{ij})||p_{\phi}(\mathbf{z}_{ij}|\mathcal{G}_i, \widetilde{\mathcal{G}}^D_j)).
\end{align}
For the intractable $p_{\phi}(\mathbf{z}_{ij}|\mathcal{G}_i, \widetilde{\mathcal{G}}^D_j)$, to make the variational inference optimizable, we introduce the reparameterization trick to derive ${p_{\phi}(\mathbf{z}_{ij}|\mathcal{G}_i, \widetilde{\mathcal{G}}^D_j)} = B(\mathbf{p}_{ij})$, where $B(\cdot)$ means the Bernoulli distribution based on the probability vector denoted as $\mathbf{p}_{ij}$. Based on this process, the VGDA can be fulfilled, where the details are described in Section~\ref{sec:VGDA} including the learning process of $\mathbf{p}_{ij}$. $q_{\phi}(\mathbf{z}_{ij}) = B(\widehat{\mathbf{p}}_{ij})$, $\widehat{\mathbf{p}}_{ij}$  means the expected probability.

By applying the reparameterization trick, the whole framework can be optimized through end-to-end back-propagation according to Eqn.~(\ref{MI_1}). Specifically, two terms are involved in Eqn.~(\ref{MI_1}), i.e. the first term (further denoted as $L_y$) constraints the accuracy for the classification task, and the second term (further denoted as $L_{KL}$)  minimizes the distribution difference between $q_{\phi}(\mathcal{Z})$ and $p_{\phi}(\mathcal{Z}|\mathcal{G}, \mathcal{D})$. Moreover, to better balance the influence of the two terms, we further introduce a trade-off coefficient $\beta$ to transform the whole optimization objective in Eqn.~(\ref{MI_1}) as follows:
\begin{align}
	\label{loss}
	\nonumber L & = L_{y} - \beta L_{KL} \\
	\nonumber          & = E_\mathbf{y|\mathcal{G},\widetilde{\mathcal{D}}}( E_{q_{\phi}(\mathcal{Z})} (\log p_{\Psi}(\mathbf{y}|\mathcal{G}, \mathcal{D})) \\  
	&  ~~~~~~~~~~~~~~~~~~~ - \beta D_{KL}(q_{\phi}(\mathcal{Z})||p_{\phi}(\mathcal{Z}|\mathcal{G}, \widetilde{\mathcal{D}})).   
\end{align}

\subsection{Variational Graph Dictionary Adaptation}\label{sec:VGDA}
We use $\mathcal{G}_i = (\mathcal{V}_i, \mathbf{X}_i, \mathbf{A}_i)~\{\mathbf{X}_i\in\mathbb{R}^{n\times d}, \mathbf{A}_i\in\mathbb{R}^{n\times n}\}$ to denote the $i$-th input graph, and $\widetilde{\mathcal{G}}^D_j = (\widetilde{\mathcal{V}}^D_j, \widetilde{\mathbf{X}}^D_j, \widetilde{\mathbf{A}}^D_j)~\{\widetilde{\mathbf{X}}^D_j\in\mathbb{R}^{n_d\times d}, \widetilde{\mathbf{A}}^D_j\in\mathbb{R}^{n_d\times n_d}\}$ to denote the $j$-th dictionary key in the BGD $\widetilde{\mathcal{D}}$. After GCNN for primary encoding, the learned features corresponding to $\mathbf{X}_i$ and $\widetilde{\mathbf{X}}^D_j$ are denoted as $\mathbf{F}_i$ and $\widetilde{\mathbf{F}}^D_j$. For each input graph, the VGDA aims to learn a more suitable AGD denoted as $\mathcal{D}$ from the BGD $\widetilde{\mathcal{D}}$. This is done by learning Bernoulli sampling factors that adaptively select substructures from the BGD. Formally, given $\mathcal{G}_i$ and $\widetilde{\mathcal{G}}^D_j$, together with the sampling factor $\mathbf{z}_{ij}$, the selection function for generating the corresponding graph $\mathcal{G}^{D}_{ij}\in\mathbb{R}^{n \times d}$ in $\mathcal{D}$ can be denoted as:
\begin{align}
	\nonumber &\mathcal{G}_{ij}^D=s(\tilde{\mathcal{G}}_{ij}^D,\mathbf{z}_{ij}), 
	&\text{s.t.}~\mathbf{z}_{ij}\sim B(\mathbf{p}_{ij}), \mathbf{p}_{ij} = g_{\phi}(\mathbf{F}_i, \tilde{\mathbf{F}}_j).
\end{align}

Each element of the vector $\mathbf{z}_{ij}$ corresponds to one node in the key graph. Specifically, $\mathbf{z}_{ij}$ conforms to the Bernoulli distribution based on the probability vector $\mathbf{p}_{ij}$, and the elements in $\mathbf{z}_{ij}$ are either 0 or 1.  Based on the selection function $s(\cdot)$ with $\mathbf{z}_{ij}$, a certain number of nodes, together with the corresponding edges, can be selected (1 for selection and 0 for discarding) from $\widetilde{\mathcal{G}}^D_j$. $g_{\phi}(\cdot)$ is the function to learn the sampling factor $\mathbf{z}_{ij}$ with the parameters denoted as $\phi$. 

Here, a crucial step is to derive $\mathbf{z}_{ij}$ based on $\mathbf{F}_i$ and $\mathbf{F}^D_j$. Rather than learning $\mathbf{z}_{ij}$ in a deterministic way, e.g. constructing networks with the attention mechanism, we infer $\mathbf{z}_{ij}$ through a probabilistic manner. Compared to the deterministic way, this probabilistic learning leads to more abundant sampling situations conforming to a predefined distribution. However, it's rather non-trivial to infer $\mathbf{z}_{ij}$ through the Bayes rule $p(\mathbf{z}_{ij}|\mathbf{F}_i,\mathbf{F}^D_j) = p(\mathbf{z}_{ij})p(\mathbf{F}_i,\mathbf{F}^D_j|\mathbf{z}_{ij})/p(\mathbf{F}_i,\mathbf{F}^D_j)$ due to the intractability. Hence, we resort to the variational inference to approximate the intractable true posterior $p(\mathbf{z}_{ij}|\mathbf{F}_i,\mathbf{F}^D_j)$ with $q_{\phi}(\mathbf{z}_{ij}|\mathbf{F}_i,\mathbf{F}^D_j)$ and meanwhile constrain the KL-divergence $D_{KL}(q_{\phi}(\mathbf{z}_{ij}|\mathbf{F}_i,\mathbf{F}^D_j)||p(\mathbf{z}_{ij}|\mathbf{F}_i,\mathbf{F}^D_j))$ between them, as described in Section~\ref{sec:MI}. For the reparameterization trick, we derive ${p(\mathbf{z}_{ij}|\mathbf{F}_i,\mathbf{F}^D_j)} = B(\mathbf{p}_{ij})$ by
\begin{align}\label{RePar}
	\mathbf{p}_{ij}= \sigma_r(cos(\mathbf{F}_i,\mathbf{F}^D_j)\mathbf{w}_r).
\end{align} 
$\mathbf{w}_r\in\mathbb{R}^{n}$ is the projection vector, $\sigma_r(\cdot)$ is the sigmoid function, $cos(\mathbf{F}_i,\mathbf{F}^D_j)$ is the correlation matrix where each element represents the cosine similarity between two vectors from $\mathbf{F}_i$ and $\mathbf{F}^D_j$, respectively.

As a result, for the input $\mathbf{F}_i$ and BGD $\widetilde{\mathcal{D}}$, the corresponding AGD $\mathcal{D}_i = \{\mathcal{G}^{D}_{i1}, \mathcal{G}_{i2}^{D}, \cdots, \mathcal{G}_{iK}^{D}\}$ can be obtained.

\subsection{Multi-sensitivity Wasserstein Embedding}\label{sec:MS-WE}

In the MS-WE module, we use the Wasserstein metric for cross-graph correlation across input graph and adapted graph keys. Moreover, the multi-sensitivity regularization is introduced to improve the sensitivity to structure variations. Formally, given the input graph $\mathcal{G}_i$, the cross-graph correlation with the adapted graph key $\mathcal{G}_{j}^{D}$ can be formularized as:      
\begin{align}
	\label{W_query} h^{\lambda}_{ij} = W_{\lambda}(\mathbf{F}_i, \mathbf{F}_{ij}^{D}) = \langle\mathbf{T}^{\lambda}_{ij}, \mathbf{M}_{ij}\rangle.
\end{align}
Each element in $\mathbf{M}_{ij}$ calculates pair-wise squared Euclidean distances between cross-graph nodes from $\mathcal{G}_i$ and $\mathcal{G}_{ij}^{D}$, respectively. $\langle\mathbf{A},\mathbf{B}\rangle = tr(\mathbf{A}^{\intercal}\mathbf{B})$. $\mathbf{T}^{\lambda}_{ij}$ represents the OT matrix between $\mathbf{F}_i$ and $\mathbf{F}_{ij}^{D}$ based on $\mathbf{M}_{ij}$, and the OT problem can be solved by the Sinkhorn's fixed point iterations:
\begin{align}
	\label{solution}
	\mathbf{T}^{\lambda}_{ij} =  \mathbf{u}_{ij}\textbf{1}_{N_1} \odot \mathbf{K}_{ij} \odot \mathbf{1}_{N_2} \mathbf{v}^{\intercal}_{ij},~s.t.~ \mathbf{K}_{ij} = e^{-\lambda \mathbf{M}_{ij}}.
\end{align}
$\mathbf{u}_{ij}$ and $\mathbf{v}_{ij}$ are initialized as all-1 vectors and kept updating during the Sinkhorn iteration. More details can be found in~ \cite{cuturi2013sinkhorn}. Hence, given the AGD $\mathcal{D}_i = \{\mathcal{G}^{D}_{i1}, \mathcal{G}_{i2}^{D}, \cdots, \mathcal{G}_{iK}^{D}\}$, the corresponding embedding $\mathbf{h}^{\lambda}_{i}= [h^{\lambda}_{i1}, \cdots, h^{\lambda}_{iK}]$ can be obtained for $\mathcal{G}_i$.

Specifically, the regulation $\lambda$ controls the sensitivity of local information between nodes across two graphs. The larger value of $\lambda$, the less sensitive to the local correlation across graphs. As large structural variation exists among graphs, one unified $\lambda$ may not well handle it for all graph samples. To address this issue, we conduct the MS-WE through two steps: (1) to calculate multi-sensitivity (e.g. C-sensitivity) embeddings denoted as $[\mathbf{h}^{\lambda_1}_{i}, \cdots, \mathbf{h}^{\lambda_C}_{i}]$ based on Eqn.~(\ref{W_query}); (2) aggregating multi-sensitivity embeddings with the attention mechanism:
\begin{align} \label{M_AT}
	\widehat{\mathbf{h}_i} = \sum_{j=1}^C \alpha_j \mathbf{h}^{\lambda_j}_{i},~s.t.~ \alpha_j = \frac{e^{(\mathbf{h}^{\lambda_j}_{i})^{\intercal}\mathbf{w}_m}}{\sum_{l=1}^{C}e^{(\mathbf{h}^{\lambda_l}_{i})^{\intercal}\mathbf{w}_m}}.
\end{align}
Specifically, $\mathbf{w}_m$ denotes the projection parameter for the attention mechanism.

Besides, the computational complexity and runtime analysis of both the VGDA and MS-WE operations can be found in the Section~\ref{sec:complexity}.

\section{Experiments}
In this section, we first introduce the public datasets used to evaluate our model, as well as describe the implementation details. We then compare the proposed SS-GDE model with multiple state-of-the-art methods. Finally, we conduct ablation analysis to dissect the SS-GDE.

\subsection{Datasets}

To comprehensively evaluate the effectiveness of SS-GDE, we conduct experiments on six widely used public datasets in the graph classification task. These datasets can be divided into two categories: bioinformatics datasets (MUTAG ~\cite{debnath1991structure}, PTC~\cite{helma2001predictive}, PROTEINS~\cite{borgwardt2005protein}) and social network datasets (COLLAB, IMDB-BINARY and IMDB-MULTI~\cite{yanardag2015deep}). The corresponding summary can be found in Table~\ref{dataset}. 

\textbf{Bioinformatics datasets.} MUTAG is a nitro compounds dataset containing 188 samples with seven discrete node labels. These samples are divided into two classes. PROTEINS comprises 1113 protein structures of secondary structure elements (SSEs) with three discrete node labels. PTC consists of 344 chemical compound networks divided into two categories, showing carcinogenicity for male and female rats. Moreover, each node is annotated with 19 labels.

\textbf{Social Network Datasets.} IMDB-BINARY and IMDB-MULTI are both movie collaboration datasets derived from IMDB, where each graph represents a movie with nodes corresponding to actors/actresses. If two actors appear in the same film, there will be an edge between their corresponding nodes. BINARY and MULTI stand for the number of classes. COLLAB is a scientific dataset where each graph represents a collaborative network between an affiliated researcher and other researchers from 3 physical domains, labeled as the researcher's physical field.

\begin{table}[!t]
	\centering
	\caption{Summary of graph datasets.}
	\label{dataset}
	\scalebox{0.8}{
		\begin{tabular}{cccccc}
			\toprule
			Datasets    & N.Graphs & Avg.Nodes & Avg.Edges & labels   & classes\\
			\midrule
			MUTAG       & 188        & 17.9          & 19.8           & 7       &2    \\
			PTC         & 344        & 25.6          & 14.7           & 19      &2      \\
			PROTEINS    & 1113       & 39.1          & 72.8           & 3       &2    \\
			IMDB-BINARY & 1000       & 19.8          & 96.5           & -       &2      \\
			IMDB-MULTI  & 1500       & 13.0          & 65.9          & -       &3      \\
			COLLAB      & 5000       & 74.5          & 2457.8          & -       &3      \\
			\bottomrule
	\end{tabular}}
\end{table}

\subsection{Experiment Setup} \label{sec:EXsetup}
\textbf{Implementation Details.} For input graph initialization, each node of the input graph is described with a one-hot vector according to its node label, and the edges are set the same as those pre-defined in the datasets.  For the base graph dictionary construction, multiple groups of graph keys, where each group corresponds to a fixed number of samples from each class, are randomly selected from the training set. In this experiment, the fixed number of keys is set to 14 for all datasets. To learn primary expression with GCNN, two encoders with the same structure, i.e. the three-layer GCNs, are employed for input samples and the base graph dictionary, respectively. Specifically, the output dimensions of the three layers are 256,128,32. In the MS-WE module, 8-sensitivity regularization is employed. Then, the output embeddings of the MS-WE module further pass two fully connected layers for classification. 

In the training stage, the framework is trained for 500 epochs with a learning rate of 0.001 and the weight decay of $10^{-4}$. In this process, almost all the parameters in the framework, together with the base graph dictionary,  are optimized through backpropagation by using the  Adam optimizer. One exception is the three-layer GCN that corresponds to the base dictionary. Specifically, its parameters are optimized according to the three-layer GCN of the inputs through the momentum mechanism. The momentum coefficient is set to 0.999. The trade-off parameter $\beta$ in Eqn.~(\ref{loss}) is set to 0.001, while the pre-defined probability vector $\widehat{\mathbf{p}}_{ij}$ in Eqn.~(\ref{KL}) is set to all-0.5 vectors. 

\textbf{Protocol.} According to the previous literature, the same 10-fold cross-validation protocol is strictly followed to evaluate the classification performance of our proposed method for a fair comparison. Specifically, We randomly split the dataset into ten sections, where nine sections are the training set, and the remaining one section is the testing set. We use the average accuracy and stand deviation of the ten folds as the final reported performance.

\subsection{Experiment Results}

\begin{table*}[h]
	\centering
	\renewcommand{\arraystretch}{1.2}
	\caption{Comparsion with the state-of-the-art-methods. The best results are in bold and $\ast$ indicates the second highest performance.}
	\scalebox{1.0}{
		\begin{tabular}[width=1linewidth]{c|cccccc}
			\hline
			Datasets & MUTAG      & PTC        & PROTEINS   & IMDB-BINARY & IMDB-MULTI & COLLAB     \\ \hline
			GK~\cite{shervashidze2009efficient}       & 81.66$\pm$2.11 & 57.26$\pm$1.41 & 71.67$\pm$0.55 & 65.87$\pm$0.98  & 43.89$\pm$0.38 & 72.84$\pm$0.28 \\
			DGK~\cite{yanardag2015deep}      & 82.66$\pm$1.45 & 57.32$\pm$1.13 & 71.68$\pm$0.50 & 66.96$\pm$0.56  & 44.55$\pm$0.52 & 73.09$\pm$0.25 \\
			WL~\cite{shervashidze2011weisfeiler}       & 82.72$\pm$3.00 & 56.97$\pm$2.01 & 73.70$\pm$0.50 & 72.86$\pm$0.76  & 50.55$\pm$0.55 & 79.02$\pm$1.77 \\
			PSCN~\cite{niepert2016learning}     & 92.63$\pm$4.21 & 62.29$\pm$5.68 & 75.89$\pm$2.76 & 71.00$\pm$2.29  & 45.23$\pm$2.84 & 72.60$\pm$2.15 \\
			NgramCNN~\cite{luo2017deep} & $94.99\pm5.63^\ast$ & 68.57$\pm$1.72 & 75.96$\pm$2.98	& 71.66$\pm$2.71  & 50.66$\pm$4.10 & -          \\
			GCN~\cite{kipf2016semi}      & 87.20$\pm$5.11 & -          & 75.65$\pm$3.24 & 73.30$\pm$5.29  & 51.20$\pm$5.13 & 81.72$\pm$1.64 \\
			GIN-0~\cite{xu2018powerful}    & 89.40$\pm$5.60 & 64.60$\pm$7.00 & 76.20$\pm$2.80 & 75.10$\pm$5.10  & 52.30$\pm$2.80 & 80.20$\pm$1.90 \\
			GNTK~\cite{du2019graph}     & 90.00$\pm$8.50 & 67.90$\pm$6.90 & 75.60$\pm$4.20 & 76.90$\pm$3.60  & 52.80$\pm$4.60 & $\textbf{83.60}\pm\textbf{1.22}$ \\
			PPGN~\cite{maron2019provably}     & 90.55$\pm$8.70 & 66.17$\pm$6.54 & 77.20$\pm$4.73 & 73.00$\pm$5.77  & 50.46$\pm$3.59 & 81.38$\pm$1.42 \\
			U2GNN~\cite{nguyen2022universal}    & 89.97$\pm$3.65 & 69.63$\pm$3.60 & $78.53\pm4.07^\ast$ & 77.04$\pm$3.45  & $53.60\pm3.53^\ast$ & 77.84$\pm$1.48 \\
			SLIM~\cite{slim_aaai22}     & 83.28$\pm$3.36 & $72.41\pm6.92^\ast$ & 77.47$\pm$4.34 & 77.23$\pm$2.12  & 53.38$\pm$4.02 & 78.22$\pm$2.02 \\
			CAL~\cite{cal_kdd22}      & 89.24$\pm$8.72 & -          & 76.28$\pm$3.65 & 74.40$\pm$4.55  & 52.13$\pm$2.96 & 82.08$\pm$2.40 \\
			HRN~\cite{hrn_ijcai22}      & 90.4$\pm$8.9   & 65.7$\pm$6.4   & -          & 77.5$\pm$4.3    & 52.8$\pm$2.7   & 81.80$\pm$1.2   \\
			GLA~\cite{yue2022label}      & 91.05$\pm$0.86 & -              & 77.45$\pm$0.38 & -        &-       & 81.54$\pm$0.14        \\
			OGDL~\cite{vincent2021online}      & 85.84$\pm$6.86 & 58.45$\pm$7.73 & 74.59$\pm$4.95 & 72.06$\pm$4.09 &50.64$\pm$4.41 & - \\
			WGDL~\cite{Zhang2021DeepWG}     & 94.68$\pm$2.63 & 70.89$\pm$5.15 & 77.29$\pm$2.91 & $79.70\pm3.59^\ast$  & 53.45$\pm$4.96 & 80.50$\pm$1.17  \\ \hline
			SS-GDE   & $\textbf{96.78}\pm\textbf{2.77}$ & $\textbf{72.96}\pm\textbf{6.39}$ & $\textbf{80.42}\pm\textbf{4.29}$ & $\textbf{80.70}\pm\textbf{4.72}$  & $\textbf{55.60}\pm\textbf{3.54}$& $82.36\pm1.44^\ast$    \\ \hline          
		\end{tabular}
		\label{results}
	}
	
\end{table*}

We compare the SS-GDE framework with a range of state-of-the-art methods as follows:
(1) graph kernel-based methods: GK~\cite{shervashidze2009efficient}, DGK~\cite{yanardag2015deep}, WL~\cite{shervashidze2011weisfeiler}, (2) Neural network based methods: PSCN~\cite{niepert2016learning}, GCN~\cite{kipf2016semi}, NgramCNN~\cite{luo2017deep}, HRN~\cite{hrn_ijcai22}, GIN-0~\cite{xu2018powerful}, U2GNN~\cite{nguyen2022universal},  GNTK~\cite{du2019graph}, PPGN~\cite{maron2019provably}, SLIM~\cite{slim_aaai22}, CAL~\cite{cal_kdd22}, GLA~\cite{yue2022label}, OGDL~\cite{vincent2021online}, and WGDL~\cite{Zhang2021DeepWG}. {These works are published in recent years and follow exactly the same evaluation protocol on the same datasets with our work.
	
Table~\ref{results} shows the experimental results of SS-GDE and the comparison with other approaches on the six public datasets. We have the following observations:
Generally, the performance of graph kernel-based methods (GK, DGK, WL) is usually lower than those GNN-based methods. This may be attributed to the low expression power caused by the usually employed hand-crafted features, and the limited feature learning ability of the two-stage learning process instead of the global optimization. Among the graph kernel-based methods, WL achieves relatively high performance on most datasets with the average performance gain of $3\%$ over GK and DGK,  as it defines a family of efficient kernels based on Weisfeiler-Lehman sequences of graphs. However, its performance is still lower than those of GNNs that stack neural network layers into a deep architecture. 

GNNs, especially GCN variants involved methods, achieve good experimental for the classification task. Specifically, several recent works improve GCN by either developing the pooling algorithm or introducing the attention mechanism for structure modeling, which further promotes the performance. According to Table~\ref{results}, our SS-GDE model achieves the state-of-the-art performances on five public datasets, and also a comparable performance on the left COLLAB dataset. This demonstrates the robustness of our framework against graph variation. In contrast, previous methods cannot perform well on all the six datasets, while the second highest performances on different datasets are obtained by six different methods. Compared with the GNTK on the other five datasets, the average performance gain of more than $4\%$ is obtained by our SS-GDE. Moreover, compared to the WGDL using a naive fixed graph dictionary and single-regularization in cross-graph modeling, the improvement of 2$\%$ on average is obtained by the SS-GDE. All above verify the effectiveness of our framework.

\subsection{Ablation Study}
As our SS-GDE framework has achieved superior performance compared to existing state-of-the-art methods, it is also meaningful to conduct additional experiments to make clear how each module promotes the classification task as well as the sensitivity of the hyper-parameters in our framework. We conduct the following ablation study.
\begin{figure*}[!htb]
	\centering
	\subfigure[Trade-off coefficient $\beta$ in Eqn.(\ref{loss})]{
		\includegraphics[width=0.23\linewidth]{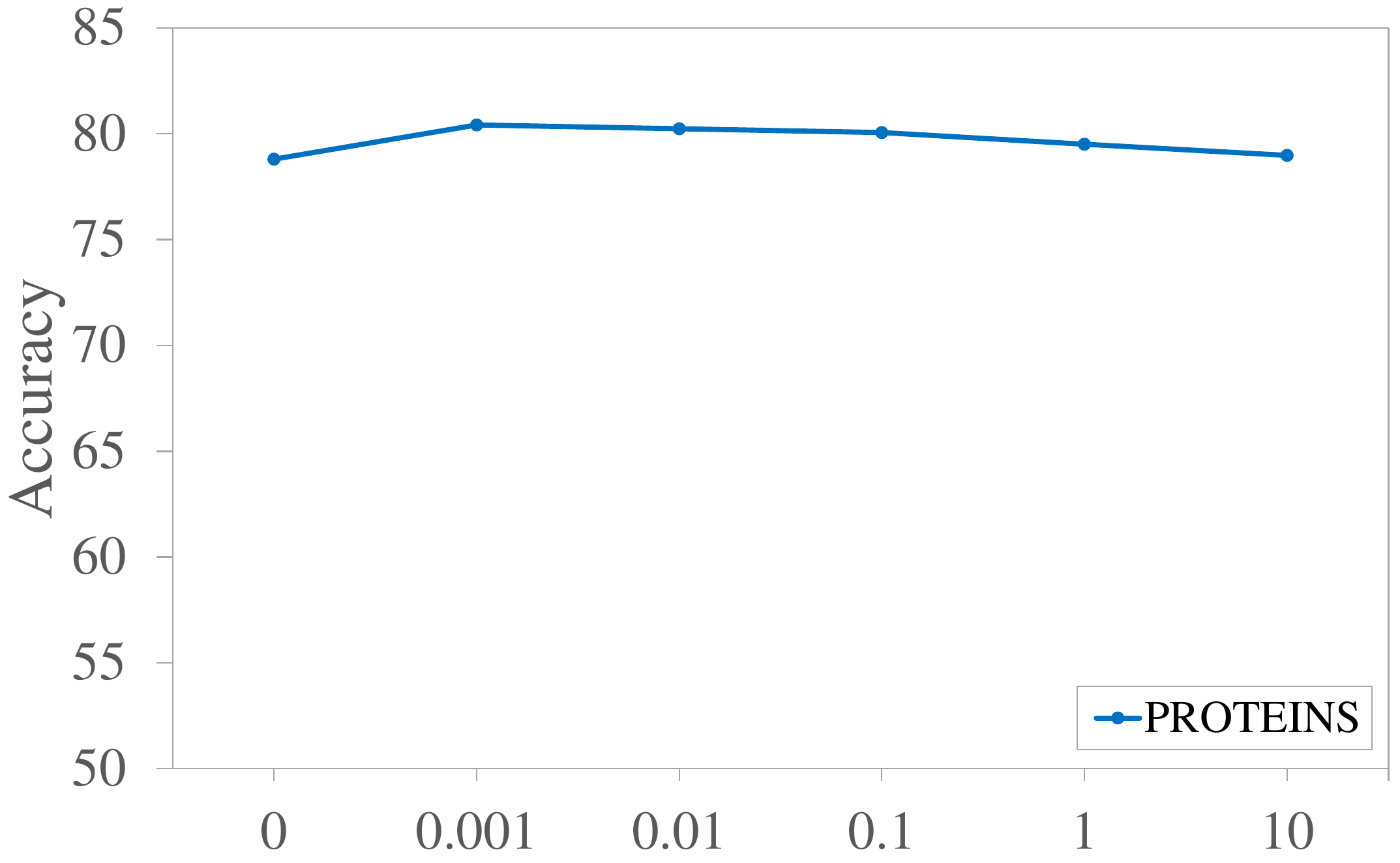}
		\label{X}
	}
	\subfigure[Expected probability $\widehat{\mathbf{p}}_{ij}$]{
		\includegraphics[width=0.23\linewidth]{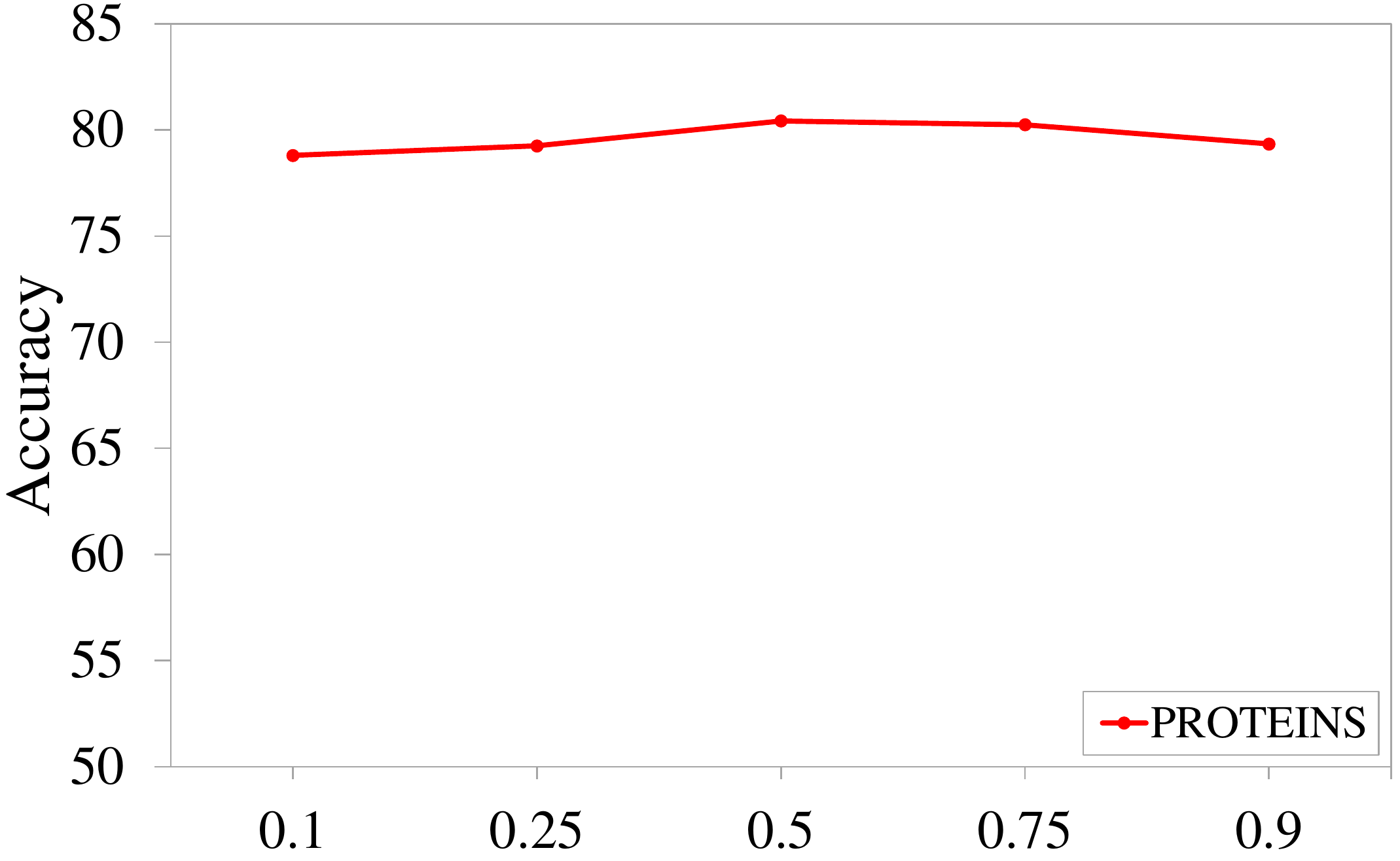}
		\label{Y}
	}
	\subfigure[Number of $\lambda$]{
		\includegraphics[width=0.23\linewidth]{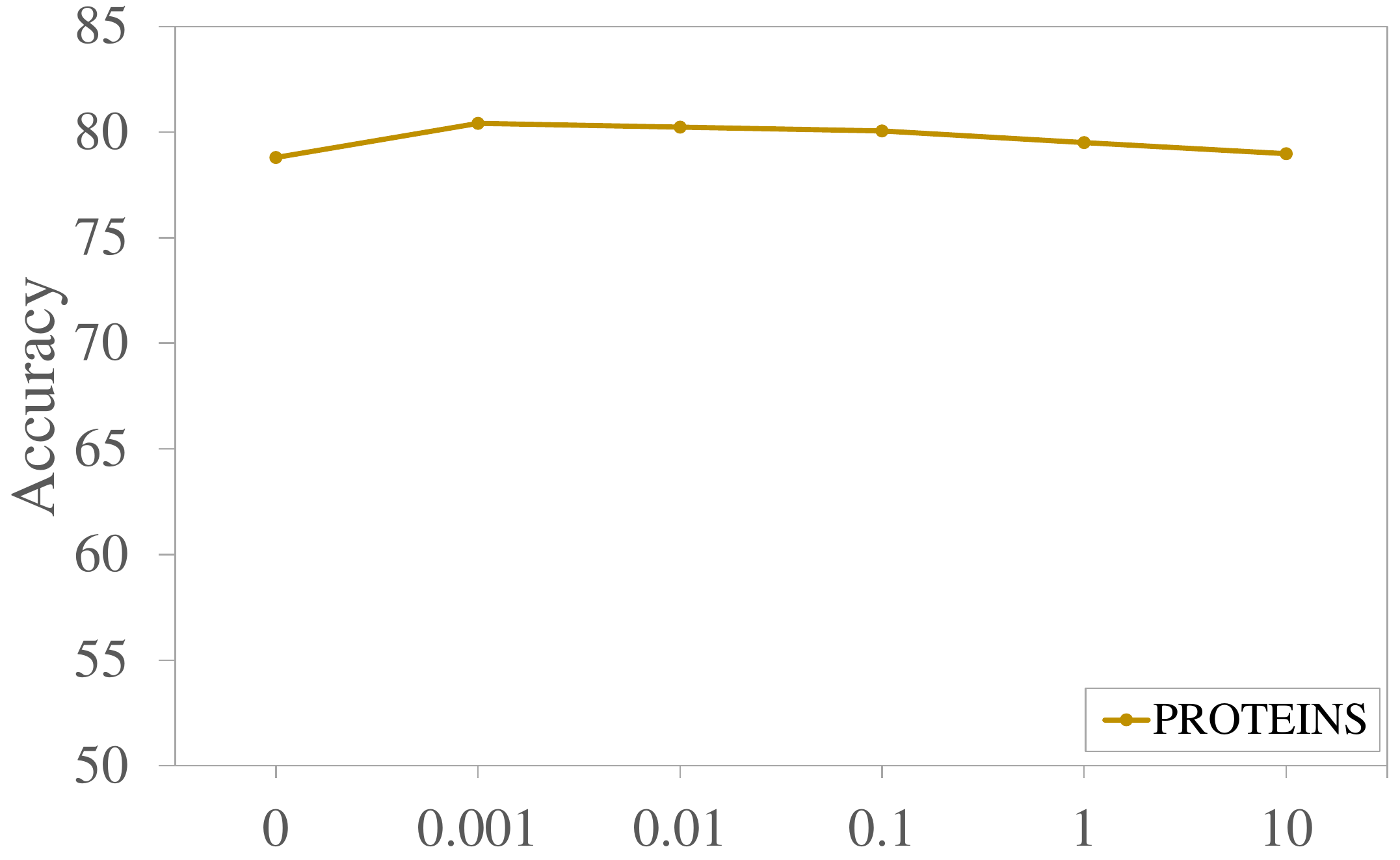}
		\label{Z}}
	\subfigure[Number of graph keys K]{
		\includegraphics[width=0.23\linewidth]{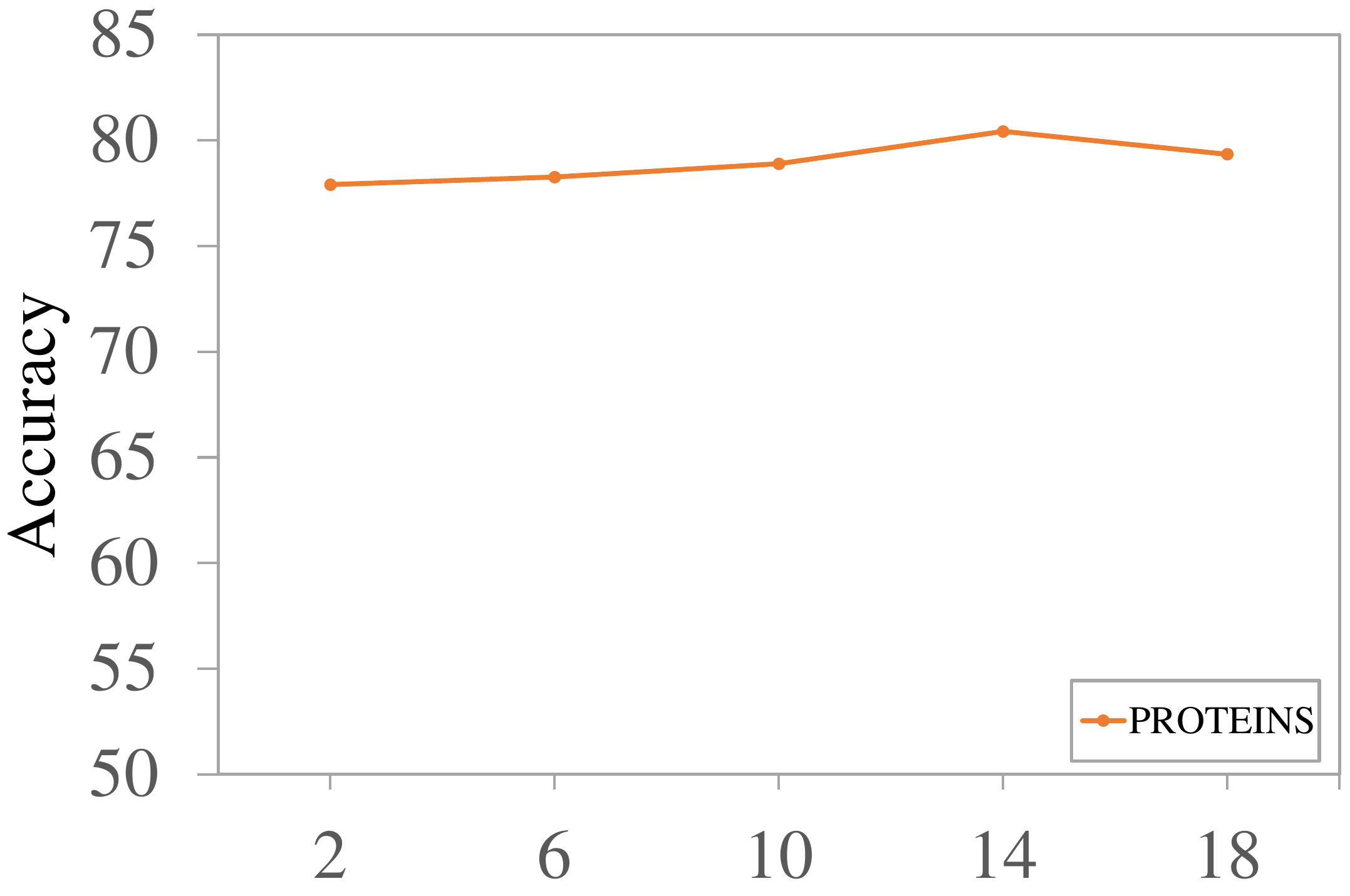}
	}
	\caption{The graph classification accuracy curves on the PROTEINS dataset with different trade-off coefficients (a), expected probabilities (b), $\lambda$ numbers (c), and graph key numbers (d).}
	\label{ablation}
\end{figure*}



\noindent \textbf{(i) To make clear the benefit of our designed VGDA module,} we just remove the VGDA module from the SS-GDE (i.e. ``BaseLine$+$MS-WE'' in Table~\ref{ablation_module}), and compared the performance.

\noindent \textbf{(ii) To verify the effectiveness of the MS-WE,} we remove the MS-WE module from the SS-GDE framework, resulting in the variant (named ``BaseLine$+$VGDA'' in Table~\ref{ablation_module}).

\noindent \textbf{(iii) The influence of the parameter $\beta$ in Eqn.~(\ref{loss}).} $\beta$ plays an essential role in balancing cross-entropy loss and Kullback-Leibler divergence for framework optimization. To quantify its influence, we vary the ratio in $\lbrace 0,0.001,0.01,0.1,1,10 \rbrace$.

\noindent \textbf{(iv) The influence of pre-defined probability vector $\widehat{\mathbf{p}}_{ij}$.} $\widehat{\mathbf{p}}_{ij}$ represents the expected probability for the Bernoulli sampling  $q_{\phi}(\mathbf{z}_{ij}) = B(\widehat{\mathbf{p}}_{ij})$ in Eqn.~(\ref{KL}). For easy implementation, we set the elements in $\widehat{\mathbf{p}}_{ij}$ to be the same value. Hence, the value actually means the expected sampling ratio of nodes from the base dictionary keys. Here, we set the value in the range of $\lbrace 0.1,0.25,0.5,0.75,0.9 \rbrace$.

\noindent \textbf{(v) The influence of the sensitivity number C in the MS-WE module.} We vary the value of C in the range in [0, 12]. For each C, the regularization parameters $\{\lambda_1, \cdots, \lambda_C\}$ in Eqn.~(\ref{M_AT}) are selected uniformly from the range $\lbrace 0.001,0.005,0.01,0.05,0.1,0.5,1,3,5,10,20,100 \rbrace$. When C is 0, only the pair-wise squared Euclidean distances across graph are calculated, without the OT.

\noindent \textbf{(vi) The influence of the dictionary key number K.} We vary the value of K in the set $\lbrace 2, 6, 10, 14, 18 \rbrace$, and compared the corresponding prediction preformance.

The results are shown in Fig.~\ref{ablation} and Table~\ref{ablation_module}. We have the following observations:
		%
\begin{table}[!t]
	\centering
	\renewcommand{\arraystretch}{1.2}
	\caption{The graph classification results with different modules exploited on the PROTEINS and IMDB-MULTI datasets.}
	\scalebox{0.8}{
		\begin{tabular}[width=1linewidth]{|c|c|c|c|c|}
			\hline
			& BaseLine[47]   & BaseLine+VGDA & Baseline+MS-WE  & SS-GDE  \\  \hline
			PROTEINS     & 77.29$\pm$2.91  & 78.35$\pm$4.30 & 79.25$\pm$3.69 & $\textbf{80.42}\pm\textbf{4.29}$ \\  \hline
			IMDB-MULTI   & 53.45$\pm$4.96 & 55.20$\pm$3.97  & 54.47$\pm$4.37 & $\textbf{55.60}\pm\textbf{3.54}$ \\ \hline
	\end{tabular}}
	\label{ablation_module}
\end{table}

\begin{figure}
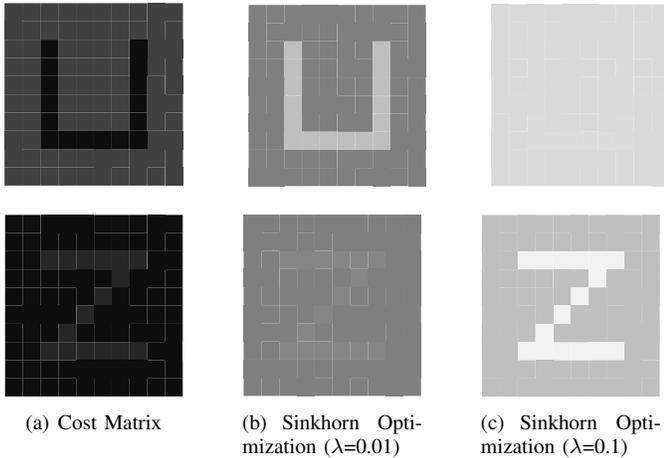

	\centering
	\subfigure{\includegraphics[width=0.27\linewidth]{./fig/ED_U.pdf}\label{ED_U}}
	\hfill
	\subfigure{\includegraphics[width=0.27\linewidth]{./fig/lambda_0.01_U.pdf}\label{U1}}
	\hfill
	\subfigure{\includegraphics[width=0.27\linewidth]{./fig/lambda_0.1_U.pdf}\label{U2}}
	\hfill
	\setcounter{subfigure}{0}
	\subfigure[Cost Matrix]{\includegraphics[width=0.27\linewidth]{./fig/ED_Z.pdf}\label{ED_Z}}
	\hfill
	\subfigure[Sinkhorn Optimization ($\lambda$=0.01)]{\includegraphics[width=0.27\linewidth]{./fig/lambda_0.01_Z.pdf}\label{Z1}}
	\hfill
	\subfigure[Sinkhorn Optimization ($\lambda$=0.1)]{\includegraphics[width=0.27\linewidth]{./fig/lambda_0.1_Z.pdf}\label{Z2}}
	\hfill
	\caption{Visualizing the influence of the sensitivity $\lambda$ on optimal transport results. Each row from the left to right shows the cross-graph cost matrix, and the corresponding Sinkhorn optimization results using different values of $\lambda$.}
	\label{necessity_MSWE}
\end{figure}

\noindent \textbf{(i)} In Table~\ref{ablation_module}, on the PROTEINS and IMDB-MULTI datasets, additionally using the VGDA effectively improves the accuracy of more than $1\%$ (``BaseLine'' vs ``BaseLine+VGDA'', ``BaseLine+MS-WE'' vs ``SS-GDE''). This verifies the importance of conducting adaptive substructure selection from the base graph dictionary. Hence, our designed VGDA module plays an important role in promoting the graph learning for classification.

\noindent \textbf{(ii)} According to Table~\ref{ablation_module}, the designed MS-WE module further promotes the graph classification performance. The average performance gain of additionally using the MS-WE (``BaseLine'' vs ``BaseLine+MS-WE'', ``BaseLine+VGDA'' vs ``SS-GDE'') on the two datasets is also more than $1\%$. This verifies the effectiveness of using the MS-WE to improve the sensitivity for cross-graph correlation.

\noindent \textbf{(iii)}  Setting the trade-off parameter $\beta$ in an appropriate range also improves the performance of our SS-GDE framework. As Fig.~\ref{ablation} (a) shows, the best performance is obtained when $\beta$ is 0.001. In this situation, the trade-off parameter well balances the first accuracy-related term and the KL-divergence in Eqn.~(\ref{loss}). When $\beta$ is set to 0, it means no distribution constraint for the VGDA. In contrast, when $\beta$ is set to 10, the KL-divergence may dominate the network optimization, while weakening the accuracy-related term.

\noindent \textbf{(iv)} According to Fig.~\ref{ablation} (b), the pre-defined probability vector $\widehat{\mathbf{p}}_{ij}$ also influences the performance, and should be set appropriately. The best performance is obtained when we set the value to 0.5. The large value of $\widehat{\mathbf{p}}_{ij}$, e.g. 1, means approximately using all the nodes in the base dictionary, significantly degrades the influence of the variational inference. In contrast, the small value of 0.1 means only about $10\%$ nodes are selected to form the AGD, which is insufficient in structure modeling.

\noindent \textbf{(v)} As it is shown in Fig.~\ref{ablation} (c), the appropriate sensitivity number can improve the sensitivity of the cross-graph correlation. The highest performance is obtained when C is set to 8. Specifically, when C is 0, low performance is achieved no optimal transport is conducted to optimize the structure in the Wasserstein space. However, not higher value of C surely achieves better performance, as too many redundant cross-graph correlations are involved. Moreover, to facilitate the intuitive understanding, we specifically visualize how different sensitivity values influence the structure capturing of the optimal transport in Fig.~\ref{necessity_MSWE}. 

\noindent \textbf{(vi)} Increasing the number of dictionary keys typically results in improved performance, as it enhances the representation capacity of the features. According to Fig.~\ref{ablation} (d), the best performance is achieved when K is set to 14. However, more keys don't surely lead to better performance. This may be attribute to the reason that excessive keys may introduce unexpected redundant information in graph correspondence measurement.

\begin{figure}[!t]
	\centering
	\includegraphics[width=1\linewidth]{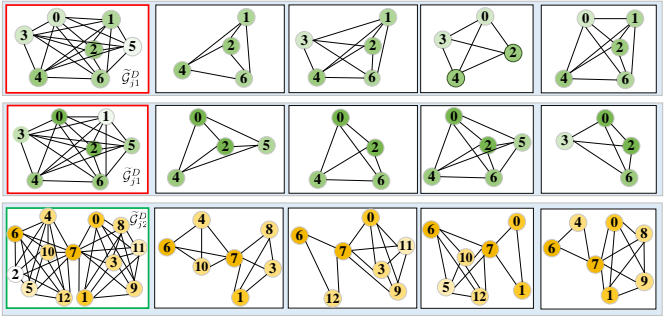}
	\caption{Visualizing the sampling examples of adapted graph keys on the IMDB-MULTI dataset. The left column shows the base graph keys in the BGD, where the darker color of each node indicates a higher sampling probability. The second to fifth columns show the different sampling situations based on the corresponding sampling probability. In the first two rows, the two graphs of the left column denote the same base graph key, but have different sampling probabilities based on different input samples. }
	\label{visual}
\end{figure}

We also visualize the sampling examples of those base graph keys in Fig.~\ref{visual}. During the framework inference, different adapted graph keys can be generated even from the sample base graph key according to the sampling probabilities. This explains why the probabilistic algorithm may generate more abundant sampling cases than the deterministic one. For different input samples, the sampling probabilities for the same graph key are also different, which indicates the adaptive sampling mechanism of our proposed VGDA module.

\subsection{The computational complexity and runtime analysis of VGDA and MS-WE operations.}\label{sec:complexity}
\begin{table}[!t]
	\centering
	\caption{The average runtime results of different methods on the PROTEINS dataset.}
	\renewcommand\arraystretch{1.5}
	\scalebox{1.0}{
		\begin{tabular}[width=1\linewidth]{|c|c|}
			\hline
			Methods     & Runtime (s) \\ 
			\hline
			BaseLine [47]  &   0.0302     \\
			\hline
			BaseLine $+$ VGDA &  0.0320       \\ 
			\hline
			BaseLine $+$ MS-WE & 0.0394        \\
			\hline
			BaseLine $+$ VGDA $+$ MS-WE (Ours)   & 0.0412 \\
			\hline
		\end{tabular}\label{Runtime}
	}
\end{table}

Let $N_1$ denote the node number of i-th input graph, $N_2$ denote the node number of j-th dictionary key, $C$ denote the number of $\lambda$, $d$ denote the dimension of hidden representations, and $t$ denote the iterations in the sinkhorn algorithm. Then, the computational complexity of VGDA and MS-WE operations are given as follows.

For the VGDA operation, it can be conducted effciently. Please see the runtime analysis (i.e. ``BaseLine$+$VGDA'' in Table~\ref{Runtime}).
The total computational complexity of VGDA is $\mathcal{O}(N_1 \cdot N_2 \cdot d + N_2)$. Specifically, it contains two main components, i.e. the computational complexity of the inference of Bernoulli sampling factor $z_{ij}$ is $\mathcal{O}(N_1 \cdot N_2 \cdot d)$, and the computational complexity of substructures selection process is $\mathcal{O}(N_2)$.

The MS-WE operation takes limited time. Compared with the baseline work[47], 0.0092s is additionally costed for inference. Please see the runtime analysis (i.e. ``BaseLine$+$MS-WE'' in Table~\ref{Runtime}). 
The total  computational complexity of MS-WE is $\mathcal{O}((d + t \cdot C + C) \cdot(N_1 \cdot N_2))$. Specifically, it contains three main components, i.e. the computational complexity of cost matrix $M_{ij}$ is $\mathcal {O}(N_1 \cdot N_2 \cdot d)$, the computational complexity of multi-sensitivity embeddings is $\mathcal{O} (t \cdot C \cdot N_1 \cdot N_2)$, and the computational complexity of aggregating embeddings is $\mathcal{O} (C \cdot N_1 \cdot N_2)$.
	
\section{Conclusion}
This paper proposed an SS-GDE framework to transform input graph into the space of graph dictionary for the graph classification task. Considering the limited expression capacity of a fixed graph dictionary for a giant amount of graphs, we propose variational graph dictionary adaptation (VGDA) to conduct individual structure selections from graph dictionary keys and generate a personalized dictionary adapted for each input graph. Besides, Bernoulli sampling is introduced to effectively choose the corresponding substructures. To increase the sensitivity and stability of cross-graph measurement, multi-sensitivity Wasserstein encoding is proposed to produce the embeddings by designing multi-scale attention on optimal transport. To optimize the proposed framework, we introduce mutual information as objective, which deduces to variational inference of adapted graph dictionary. We evaluated the SS-GDE on multiple datasets and dissected the framework with ablation analysis. The experimental results demonstrate the effectiveness and superiority of our model. 

\section*{Appendix}

In the following part, we show the detailed derivation of Eqn.~(\ref{MI_1}). 
Our optimization objective in Eqn.~(\ref{obj}) aims to maximize the mutual information between two variables, i.e. (i) the variable $\mathbf{y}$, distributed in the label space denoted as $\mathcal{Y}$; and (ii) the variable corresponding to $f_{\Psi,\phi}(\mcG, \mcD)$, further denoted as $\mathbf{s}$ here for simplification, which is distributed in the embedding space $\mathcal{S}$. Specifically, $\mathcal{S}$ is spanned by the resulted embeddings of $f_{\Psi,\phi}(\cdot)$ inferred based on the input $\mcG$ and base graph dictionary $\tilde{\mcD}$. 

Then, based on the connection between the mutual information $I(\cdot)$ and entropy $H(\cdot)$, the objective in Eqn.~(\ref{MI_1}) can be further written as:
\begin{equation}
	\nonumber  I(\mathbf{y}, \mathbf{s}) =-H(\mathbf{y}|\mathbf{s})+H(\mathbf{y}).
\end{equation}
In the optimization above, the second term $H(\mathbf{y})$ can be just omitted from the objective as it is independent from those parameters. Then, we have the following derivation:    
\begin{align}
	\nonumber                & -H(\mathbf{y}|\mathbf{s}) \\
	\nonumber                                          & =  \sum_{i} ~-H(\mathbf{y}|\mathbf{s} = S_i)p(S_i) \\
	\nonumber                                          & = \sum_{i}~p(S_i) E_{\mathbf{y}|S_i}(\log p(\mathbf{y}|\mathbf{s} = S_i)). 
\end{align}

Specifically, $p(S_i)$ means the probability of the $i$-th observation in the embedding space, and can be rationally assumed to conform the uniform distribution. As the observation $S_i$ means to be inferenced based on an input   sample $\mcG_i$ with $\tilde{\mcD}$, we further denote  $p(\mathbf{y}|\mathbf{s} = S_i)$   equally as  $p_{\Psi,\phi}(\mathbf{y}|\mcG_i, \widetilde{\mathcal{D}})$, where $\Psi,\phi$ are two parameter sets introduced in the manuscript. Assuming $L$ is the number of all observations of $\mathbf{s}$, the objective above can be further written as: 
\begin{align}
	\nonumber                     &=\sum_{i} \frac{1}{L} E_{\mathbf{y}|\mcG_i, \mcD}(\log p_{\Psi,\phi}(\mathbf{y}|\mcG_i, \widetilde{\mathcal{D}})) \\  
	\nonumber                     &=\sum_{i}  E_{\mathbf{y}|\mcG_i, \widetilde{\mathcal{D}}}(\log \int  p_{\Psi}(\mathbf{y}|\mcG_i, \widetilde{\mathcal{D}}, \mcZ)  p_{\phi}(\mcZ|\mcG_i,\widetilde{\mathcal{D}})d\mcZ)\\  
	\nonumber                     &=\sum_{i}  E_{\mathbf{y}|\mcG_i, \widetilde{\mathcal{D}}}(\log \int  p_{\Psi}(\mathbf{y}|\mcG_i,\mathcal{D})  p_{\phi}(\mcZ|\mcG_i,\widetilde{\mathcal{D}})d\mcZ)\\  
	\nonumber                     &=\sum_{i}  E_{\mathbf{y}|\mcG_i, \widetilde{\mathcal{D}}}(\log \int q_{\phi}(\mcZ|\mcG_i, \widetilde{\mathcal{D}}) p_{\Psi}(\mathbf{y}|\mcG_i, \mathcal{D}) \frac{p_{\phi}(\mcZ|\mcG_i, \widetilde{\mathcal{D}})}{q_{\phi}(\mcZ|\mcG_i, \widetilde{\mathcal{D}})}d\mcZ) 
\end{align}
According to the Jensen Inequality, the above formula can be deduced as :
\begin{align}
	\nonumber                     & \ge \sum_{i} E_{\mathbf{y}|\mcG_i, \widetilde{\mathcal{D}}}(\int q_{\phi}(\mcZ|\mcG_i, \widetilde{\mathcal{D}}) \log p_{\Psi}(\mathbf{y}|\mcG_i, \mcD) \frac{p_{\phi}(\mcZ|\mcG_i, \widetilde{\mathcal{D}})}{q_{\phi}(\mcZ|\mcG_i, \widetilde{\mathcal{D}})}d\mcZ) \\
	\nonumber                     & = \sum_{i}E_{\mathbf{y}|\mcG_i, \widetilde{\mathcal{D}}}(\int q_{\phi}(\mcZ|\mcG_i, \widetilde{\mathcal{D}}) \log p_{\Psi}(\mathbf{y}|\mcG_i, \mcD) \\
	\nonumber                     &  ~~~~~~~~~~~~~~~~~~~ ~~~~~~~~~ ~~+ q_{\phi}(\mcZ|\mcG_i, \widetilde{\mathcal{D}}) \log \frac{p_{\phi}(\mcZ|\mcG_i, \widetilde{\mathcal{D}})}{q_{\phi}(\mcZ|\mcG_i, \widetilde{\mathcal{D}})}d\mcZ) \\
	\nonumber                     & = \sum_{i} E_{\mathbf{y}|\mcG_i, \widetilde{\mathcal{D}}}(E_{q_{\phi}(\mcZ|\mcG_i, \widetilde{\mathcal{D}})} (\log p_{\Psi}(\mathbf{y}|\mcG_i, \mcD)) \\
	\nonumber                     &  ~~~~~~~~~~~~ ~~~~~~~~~~~~~~~~~~ ~- D_{KL}(q_{\phi}(\mcZ|\mcG_i, \widetilde{\mathcal{D}})||p_{\phi}(\mcZ|\mcG_i, \widetilde{\mathcal{D}}).
\end{align}

Specifically, for the expected distribution $q_{\phi}(\mcZ|\mcG_i, \widetilde{\mathcal{D}})$, the distribution is assumed independent from $\mcG_i$ and  $\widetilde{\mathcal{D}}$. Hence, $q_{\phi}(\mcZ|\mcG_i, \widetilde{\mathcal{D}})$ can be rewritten as $q_{\phi}(\mcZ)$, which completes the proof of Eqn.~(\ref{MI_1}).

\bibliographystyle{IEEEtran}
\bibliography{egbib.bib}

\begin{thebibliography}{10}
\providecommand{\url}[1]{#1}
\csname url@samestyle\endcsname
\providecommand{\newblock}{\relax}
\providecommand{\bibinfo}[2]{#2}
\providecommand{\BIBentrySTDinterwordspacing}{\spaceskip=0pt\relax}
\providecommand{\BIBentryALTinterwordstretchfactor}{4}
\providecommand{\BIBentryALTinterwordspacing}{\spaceskip=\fontdimen2\font plus
\BIBentryALTinterwordstretchfactor\fontdimen3\font minus
  \fontdimen4\font\relax}
\providecommand{\BIBforeignlanguage}[2]{{%
\expandafter\ifx\csname l@#1\endcsname\relax
\typeout{** WARNING: IEEEtran.bst: No hyphenation pattern has been}%
\typeout{** loaded for the language `#1'. Using the pattern for}%
\typeout{** the default language instead.}%
\else
\language=\csname l@#1\endcsname
\fi
#2}}
\providecommand{\BIBdecl}{\relax}
\BIBdecl

\bibitem{duvenaud2015convolutional}
D.~K. Duvenaud, D.~Maclaurin, J.~Iparraguirre, R.~Bombarell, T.~Hirzel,
  A.~Aspuru-Guzik, and R.~P. Adams, ``Convolutional networks on graphs for
  learning molecular fingerprints,'' \emph{Advances in neural information
  processing systems}, vol.~28, 2015.

\bibitem{hamilton2017inductive}
W.~Hamilton, Z.~Ying, and J.~Leskovec, ``Inductive representation learning on
  large graphs,'' \emph{Advances in neural information processing systems},
  vol.~30, 2017.

\bibitem{borgwardt2005shortest}
K.~M. Borgwardt and H.-P. Kriegel, ``Shortest-path kernels on graphs,'' in
  \emph{Fifth IEEE international conference on data mining (ICDM'05)}.\hskip
  1em plus 0.5em minus 0.4em\relax IEEE, 2005, pp. 8--pp.

\bibitem{shervashidze2009efficient}
N.~Shervashidze, S.~Vishwanathan, T.~Petri, K.~Mehlhorn, and K.~Borgwardt,
  ``Efficient graphlet kernels for large graph comparison,'' in
  \emph{Artificial intelligence and statistics}.\hskip 1em plus 0.5em minus
  0.4em\relax PMLR, 2009, pp. 488--495.

\bibitem{kipf2016semi}
T.~N. Kipf and M.~Welling, ``Semi-supervised classification with graph
  convolutional networks,'' \emph{arXiv preprint arXiv:1609.02907}, 2016.

\bibitem{velivckovic2017graph}
P.~Veli{\v{c}}kovi{\'c}, G.~Cucurull, A.~Casanova, A.~Romero, P.~Lio, and
  Y.~Bengio, ``Graph attention networks,'' \emph{arXiv preprint
  arXiv:1710.10903}, 2017.

\bibitem{xu2018powerful}
K.~Xu, W.~Hu, J.~Leskovec, and S.~Jegelka, ``How powerful are graph neural
  networks?'' \emph{arXiv preprint arXiv:1810.00826}, 2018.

\bibitem{Zhang2021DeepWG}
T.~Zhang, Y.~Wang, Z.~Cui, C.~Zhou, B.~Cui, H.~Huang, and J.~Yang, ``Deep
  wasserstein graph discriminant learning for graph classification,'' in
  \emph{AAAI}, 2021.

\bibitem{vincent2021online}
C.~Vincent-Cuaz, T.~Vayer, R.~Flamary, M.~Corneli, and N.~Courty, ``Online
  graph dictionary learning,'' in \emph{International Conference on Machine
  Learning}.\hskip 1em plus 0.5em minus 0.4em\relax PMLR, 2021, pp.
  10\,564--10\,574.

\bibitem{gartner2003graph}
T.~G{\"a}rtner, P.~Flach, and S.~Wrobel, ``On graph kernels: Hardness results
  and efficient alternatives,'' in \emph{Learning theory and kernel
  machines}.\hskip 1em plus 0.5em minus 0.4em\relax Springer, 2003, pp.
  129--143.

\bibitem{shervashidze2011weisfeiler}
N.~Shervashidze, P.~Schweitzer, E.~J. Van~Leeuwen, K.~Mehlhorn, and K.~M.
  Borgwardt, ``Weisfeiler-lehman graph kernels.'' \emph{Journal of Machine
  Learning Research}, vol.~12, no.~9, 2011.

\bibitem{luo2017deep}
Z.~Luo, L.~Liu, J.~Yin, Y.~Li, and Z.~Wu, ``Deep learning of graphs with ngram
  convolutional neural networks,'' \emph{IEEE Transactions on Knowledge and
  Data Engineering}, vol.~29, no.~10, pp. 2125--2139, 2017.

\bibitem{niepert2016learning}
M.~Niepert, M.~Ahmed, and K.~Kutzkov, ``Learning convolutional neural networks
  for graphs,'' in \emph{International conference on machine learning}.\hskip
  1em plus 0.5em minus 0.4em\relax PMLR, 2016, pp. 2014--2023.

\bibitem{lee2019self}
J.~Lee, I.~Lee, and J.~Kang, ``Self-attention graph pooling,'' in
  \emph{International conference on machine learning}.\hskip 1em plus 0.5em
  minus 0.4em\relax PMLR, 2019, pp. 3734--3743.

\bibitem{li2020graph}
M.~Li, S.~Chen, Y.~Zhang, and I.~Tsang, ``Graph cross networks with vertex
  infomax pooling,'' \emph{Advances in Neural Information Processing Systems},
  vol.~33, pp. 14\,093--14\,105, 2020.

\bibitem{nouranizadeh2021maximum}
A.~Nouranizadeh, M.~Matinkia, M.~Rahmati, and R.~Safabakhsh, ``Maximum entropy
  weighted independent set pooling for graph neural networks,'' \emph{arXiv
  preprint arXiv:2107.01410}, 2021.

\bibitem{ying2018hierarchical}
Z.~Ying, J.~You, C.~Morris, X.~Ren, W.~Hamilton, and J.~Leskovec,
  ``Hierarchical graph representation learning with differentiable pooling,''
  \emph{Advances in neural information processing systems}, vol.~31, 2018.

\bibitem{yuan2020structpool}
H.~Yuan and S.~Ji, ``Structpool: Structured graph pooling via conditional
  random fields,'' in \emph{Proceedings of the 8th International Conference on
  Learning Representations}, 2020.

\bibitem{gmt_iclr21}
J.~Baek, M.~Kang, and S.~J. Hwang, ``Accurate learning of graph representations
  with graph multiset pooling,'' in \emph{International Conference on Learning
  Representations (ICLR)}, 2021.

\bibitem{li2019semi}
J.~Li, Y.~Rong, H.~Cheng, H.~Meng, W.~Huang, and J.~Huang, ``Semi-supervised
  graph classification: A hierarchical graph perspective,'' in \emph{The World
  Wide Web Conference}, 2019, pp. 972--982.

\bibitem{becigneul2020optimal}
G.~B{\'e}cigneul, O.-E. Ganea, B.~Chen, R.~Barzilay, and T.~S. Jaakkola,
  ``Optimal transport graph neural networks,'' 2020.

\bibitem{chen2020graph}
L.~Chen, Z.~Gan, Y.~Cheng, L.~Li, L.~Carin, and J.~Liu, ``Graph optimal
  transport for cross-domain alignment,'' in \emph{International Conference on
  Machine Learning}.\hskip 1em plus 0.5em minus 0.4em\relax PMLR, 2020, pp.
  1542--1553.

\bibitem{frogner2015learning}
C.~Frogner, C.~Zhang, H.~Mobahi, M.~Araya, and T.~A. Poggio, ``Learning with a
  wasserstein loss,'' \emph{Advances in neural information processing systems},
  vol.~28, 2015.

\bibitem{titouan2019optimal}
V.~Titouan, N.~Courty, R.~Tavenard, and R.~Flamary, ``Optimal transport for
  structured data with application on graphs,'' in \emph{International
  Conference on Machine Learning}.\hskip 1em plus 0.5em minus 0.4em\relax PMLR,
  2019, pp. 6275--6284.

\bibitem{vincent2022template}
C.~Vincent-Cuaz, R.~Flamary, M.~Corneli, T.~Vayer, and N.~Courty, ``Template
  based graph neural network with optimal transport distances,'' \emph{arXiv
  preprint arXiv:2205.15733}, 2022.

\bibitem{togninalli2019wasserstein}
M.~Togninalli, E.~Ghisu, F.~Llinares-L{\'o}pez, B.~Rieck, and K.~Borgwardt,
  ``Wasserstein weisfeiler-lehman graph kernels,'' \emph{Advances in Neural
  Information Processing Systems}, vol.~32, 2019.

\bibitem{rolet2016fast}
A.~Rolet, M.~Cuturi, and G.~Peyr{\'e}, ``Fast dictionary learning with a
  smoothed wasserstein loss,'' in \emph{Artificial Intelligence and
  Statistics}.\hskip 1em plus 0.5em minus 0.4em\relax PMLR, 2016, pp. 630--638.

\bibitem{schmitz2018wasserstein}
M.~A. Schmitz, M.~Heitz, N.~Bonneel, F.~Ngole, D.~Coeurjolly, M.~Cuturi,
  G.~Peyr{\'e}, and J.-L. Starck, ``Wasserstein dictionary learning: Optimal
  transport-based unsupervised nonlinear dictionary learning,'' \emph{SIAM
  Journal on Imaging Sciences}, vol.~11, no.~1, pp. 643--678, 2018.

\bibitem{dai2021towards}
E.~Dai and S.~Wang, ``Towards self-explainable graph neural network,'' in
  \emph{Proceedings of the 30th ACM International Conference on Information \&
  Knowledge Management}, 2021, pp. 302--311.

\bibitem{lin2022orphicx}
W.~Lin, H.~Lan, H.~Wang, and B.~Li, ``Orphicx: A causality-inspired latent
  variable model for interpreting graph neural networks,'' in \emph{Proceedings
  of the IEEE/CVF Conference on Computer Vision and Pattern Recognition}, 2022,
  pp. 13\,729--13\,738.

\bibitem{luo2020parameterized}
D.~Luo, W.~Cheng, D.~Xu, W.~Yu, B.~Zong, H.~Chen, and X.~Zhang, ``Parameterized
  explainer for graph neural network,'' \emph{Advances in neural information
  processing systems}, vol.~33, pp. 19\,620--19\,631, 2020.

\bibitem{miao2022interpretable}
S.~Miao, M.~Liu, and P.~Li, ``Interpretable and generalizable graph learning
  via stochastic attention mechanism,'' in \emph{International Conference on
  Machine Learning}.\hskip 1em plus 0.5em minus 0.4em\relax PMLR, 2022, pp.
  15\,524--15\,543.

\bibitem{wu2022discovering}
Y.-X. Wu, X.~Wang, A.~Zhang, X.~He, and T.-S. Chua, ``Discovering invariant
  rationales for graph neural networks,'' \emph{arXiv preprint
  arXiv:2201.12872}, 2022.

\bibitem{ying2019gnnexplainer}
Z.~Ying, D.~Bourgeois, J.~You, M.~Zitnik, and J.~Leskovec, ``Gnnexplainer:
  Generating explanations for graph neural networks,'' \emph{Advances in neural
  information processing systems}, vol.~32, 2019.

\bibitem{yu2022improving}
J.~Yu, J.~Cao, and R.~He, ``Improving subgraph recognition with variational
  graph information bottleneck,'' in \emph{Proceedings of the IEEE/CVF
  Conference on Computer Vision and Pattern Recognition}, 2022, pp.
  19\,396--19\,405.

\bibitem{yu2020graph}
J.~Yu, T.~Xu, Y.~Rong, Y.~Bian, J.~Huang, and R.~He, ``Graph information
  bottleneck for subgraph recognition,'' \emph{arXiv preprint
  arXiv:2010.05563}, 2020.

\bibitem{cuturi2013sinkhorn}
M.~Cuturi, ``Sinkhorn distances: Lightspeed computation of optimal transport,''
  \emph{Advances in neural information processing systems}, vol.~26, 2013.

\bibitem{debnath1991structure}
A.~K. Debnath, R.~L. Lopez~de Compadre, G.~Debnath, A.~J. Shusterman, and
  C.~Hansch, ``Structure-activity relationship of mutagenic aromatic and
  heteroaromatic nitro compounds. correlation with molecular orbital energies
  and hydrophobicity,'' \emph{Journal of medicinal chemistry}, vol.~34, no.~2,
  pp. 786--797, 1991.

\bibitem{helma2001predictive}
C.~Helma, R.~D. King, S.~Kramer, and A.~Srinivasan, ``The predictive toxicology
  challenge 2000--2001,'' \emph{Bioinformatics}, vol.~17, no.~1, pp. 107--108,
  2001.

\bibitem{borgwardt2005protein}
K.~M. Borgwardt, C.~S. Ong, S.~Sch{\"o}nauer, S.~Vishwanathan, A.~J. Smola, and
  H.-P. Kriegel, ``Protein function prediction via graph kernels,''
  \emph{Bioinformatics}, vol.~21, no. suppl\_1, pp. i47--i56, 2005.

\bibitem{yanardag2015deep}
P.~Yanardag and S.~Vishwanathan, ``Deep graph kernels,'' in \emph{Proceedings
  of the 21th ACM SIGKDD international conference on knowledge discovery and
  data mining}, 2015, pp. 1365--1374.

\bibitem{du2019graph}
S.~S. Du, K.~Hou, R.~R. Salakhutdinov, B.~Poczos, R.~Wang, and K.~Xu, ``Graph
  neural tangent kernel: Fusing graph neural networks with graph kernels,''
  \emph{Advances in neural information processing systems}, vol.~32, 2019.

\bibitem{maron2019provably}
H.~Maron, H.~Ben-Hamu, H.~Serviansky, and Y.~Lipman, ``Provably powerful graph
  networks,'' \emph{Advances in neural information processing systems},
  vol.~32, 2019.

\bibitem{nguyen2022universal}
D.~Q. Nguyen, T.~D. Nguyen, and D.~Phung, ``Universal graph transformer
  self-attention networks,'' in \emph{Companion Proceedings of the Web
  Conference 2022}, 2022, pp. 193--196.

\bibitem{slim_aaai22}
Y.~Zhu, K.~Zhang, J.~Wang, H.~Ling, J.~Zhang, and H.~Zha, ``Structural
  landmarking and interaction modelling: A "slim" network for graph
  classification,'' in \emph{Proceedings of the Thirty-Sixth Conference on
  Association for the Advancement of Artificial Intelligence (AAAI)}, 2022, pp.
  9251--9259.

\bibitem{cal_kdd22}
Y.~Sui, X.~Wang, J.~Wu, M.~Lin, X.~He, and T.-S. Chua, ``Causal attention for
  interpretable and generalizable graph classification,'' in \emph{Proceedings
  of the 28th ACM SIGKDD Conference on Knowledge Discovery and Data Mining
  (KDD)}, 2022, pp. 1696--1705.

\bibitem{hrn_ijcai22}
J.~Wu, S.~Li, J.~Li, Y.~Pan, and K.~Xu, ``A simple yet effective method for
  graph classification,'' in \emph{Proceedings of the Thirty-First
  International Joint Conference on Artificial Intelligence (IJCAI)}, 2022, pp.
  3580--3586.

\bibitem{yue2022label}
H.~Yue, C.~Zhang, C.~Zhang, and H.~Liu, ``Label-invariant augmentation for
  semi-supervised graph classification,'' \emph{arXiv preprint
  arXiv:2205.09802}, 2022.

\end{thebibliography}

\begin{IEEEbiography}[{\includegraphics[width=1in,height=1.25in,clip,keepaspectratio]{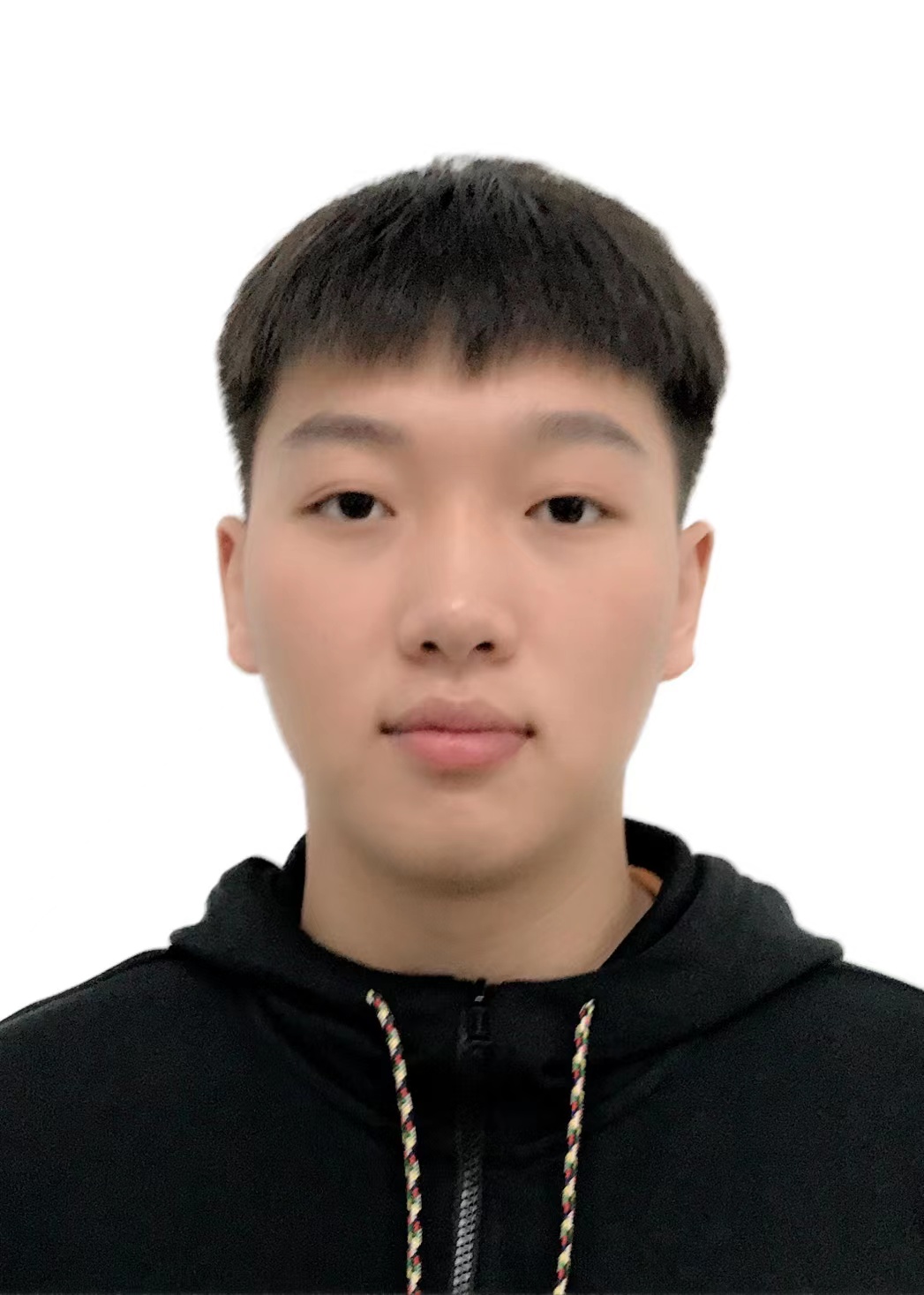}}]{Guangbu Liu} received the B.S. degree in Software Engineering from Zhongyuan University of Technology, Zhengzhou, China, in 2021. He is currently working toward the M.S. degree at the School of Computer and Engineering, Nanjing University of Science and Technology, Nanjing, 210094, China. Her current reserach interests include pattern recoginition and deep learning.
\end{IEEEbiography}

\begin{IEEEbiography}[{\includegraphics[width=1in,height=1.25in,clip,keepaspectratio]{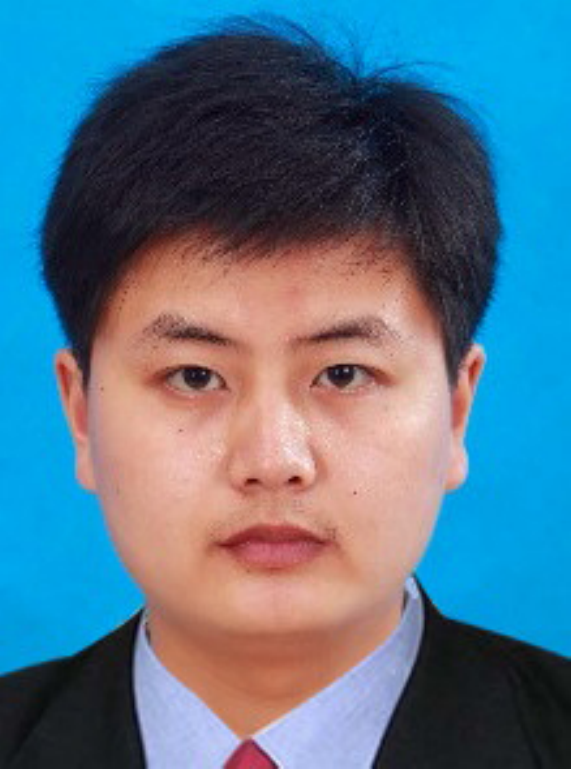}}]{Tong Zhang} received the B.S. degree from the School of Information Science and Engineering, Southeast University, Nanjing, China, in 2011, the M.S. degree from the Research Center for Learning Science, Southeast University, Nanjing, China, in 2014, and the Ph.D. degree from the School of Information Science and Engineering, Southeast University, in 2018. He is currently a Lecturer with the School of Computer Science and Engineering, Nanjing University of Science and Technology, Nanjing, China. His
research interests include pattern recognition, machine learning, and computer vision.
\end{IEEEbiography}

\begin{IEEEbiography}[{\includegraphics[width=1in,height=1.25in,clip,keepaspectratio]{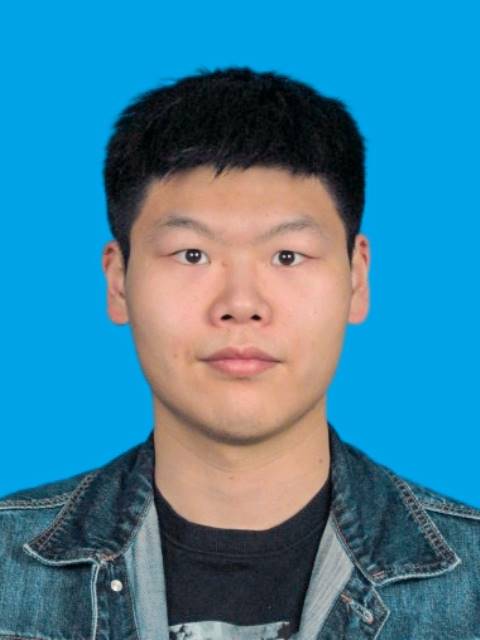}}]{Xudong Wang} received the B. S. degree from the School of Software, Xidian University, Xi’an, China, in 2018, the M. S. degree from the School of Computer Science and Technology, Xidian University, Xi’an, China. He is currently a Doctor with the School of Computer Science and Engineering, Nanjing University of Science and Technology, Nanjing, China. His research interests include pattern recognition, deep learning, graph neural network.
\end{IEEEbiography}

\begin{IEEEbiography}[{\includegraphics[width=1in,height=1.25in,clip,keepaspectratio]{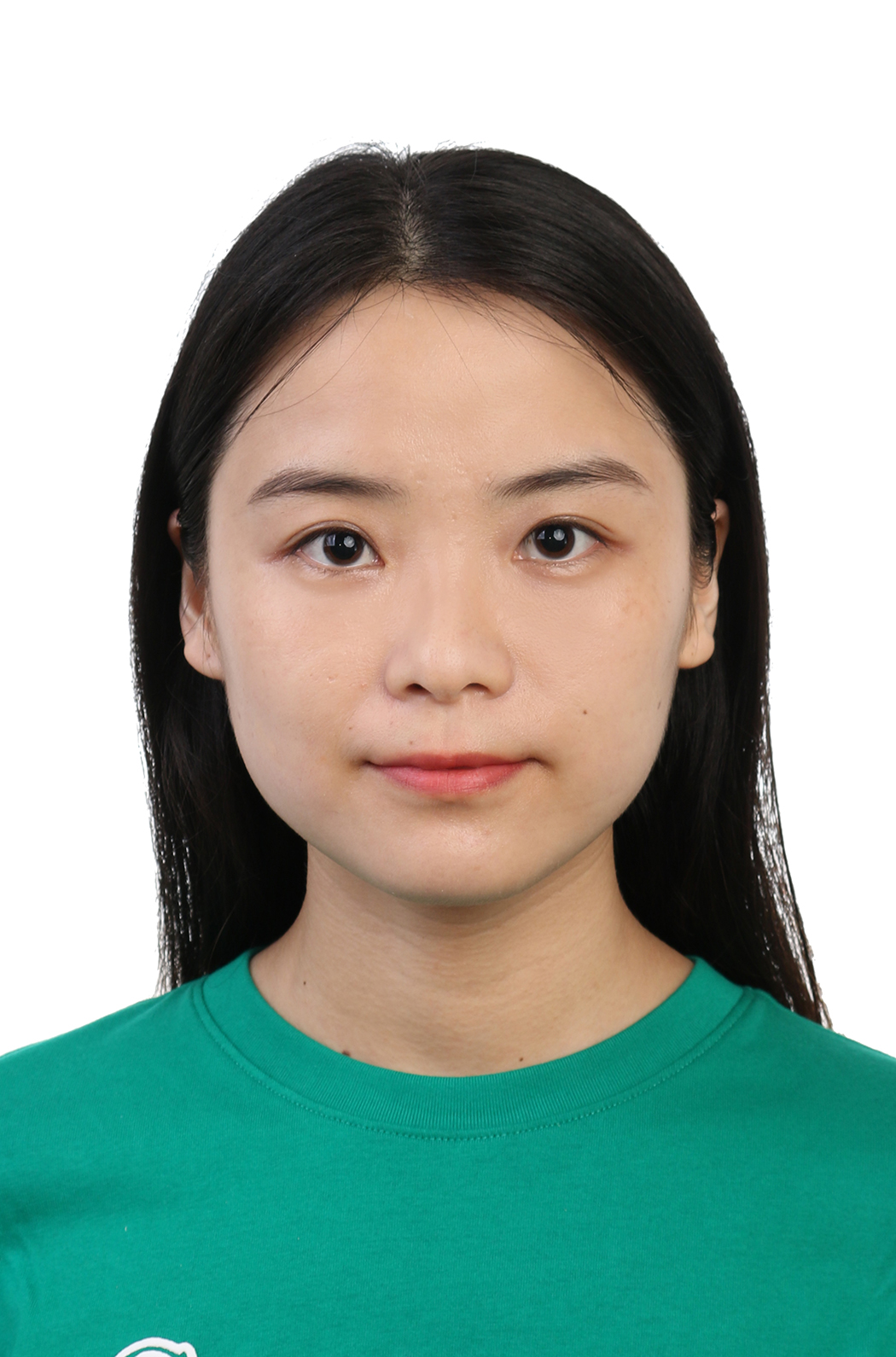}}]{Wenting Zhao} is a Ph.D. student in the School of Computer Science and Engineering from Nanjing University of Science and Technology, Nanjing, 210094. Her research interests focus on graph mining, machine leanring, protein sturucture prediction.
\end{IEEEbiography}

\begin{IEEEbiography}[{\includegraphics[width=1in,height=1.25in,clip,keepaspectratio]{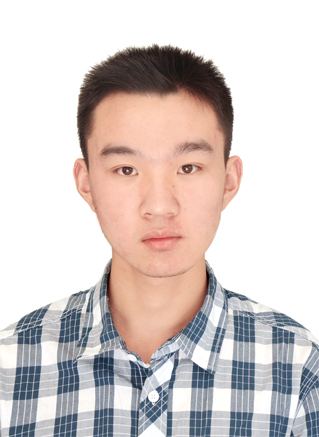}}]{Chuanwei Zhou} received the B.S. degree from the College of Elite Education, Nanjing University of Science and Technology (NUST), Jiangsu, China, in 2017.  He is currently working toward the PhD degree in Computer Science and Technology with the Nanjing University of Science and Technology, Jiangsu, China.
	His research interests include computer vision, pattern recognition, data mining and deep learning.
\end{IEEEbiography}

\begin{IEEEbiography}[{\includegraphics[width=1in,height=1.25in,clip,keepaspectratio]{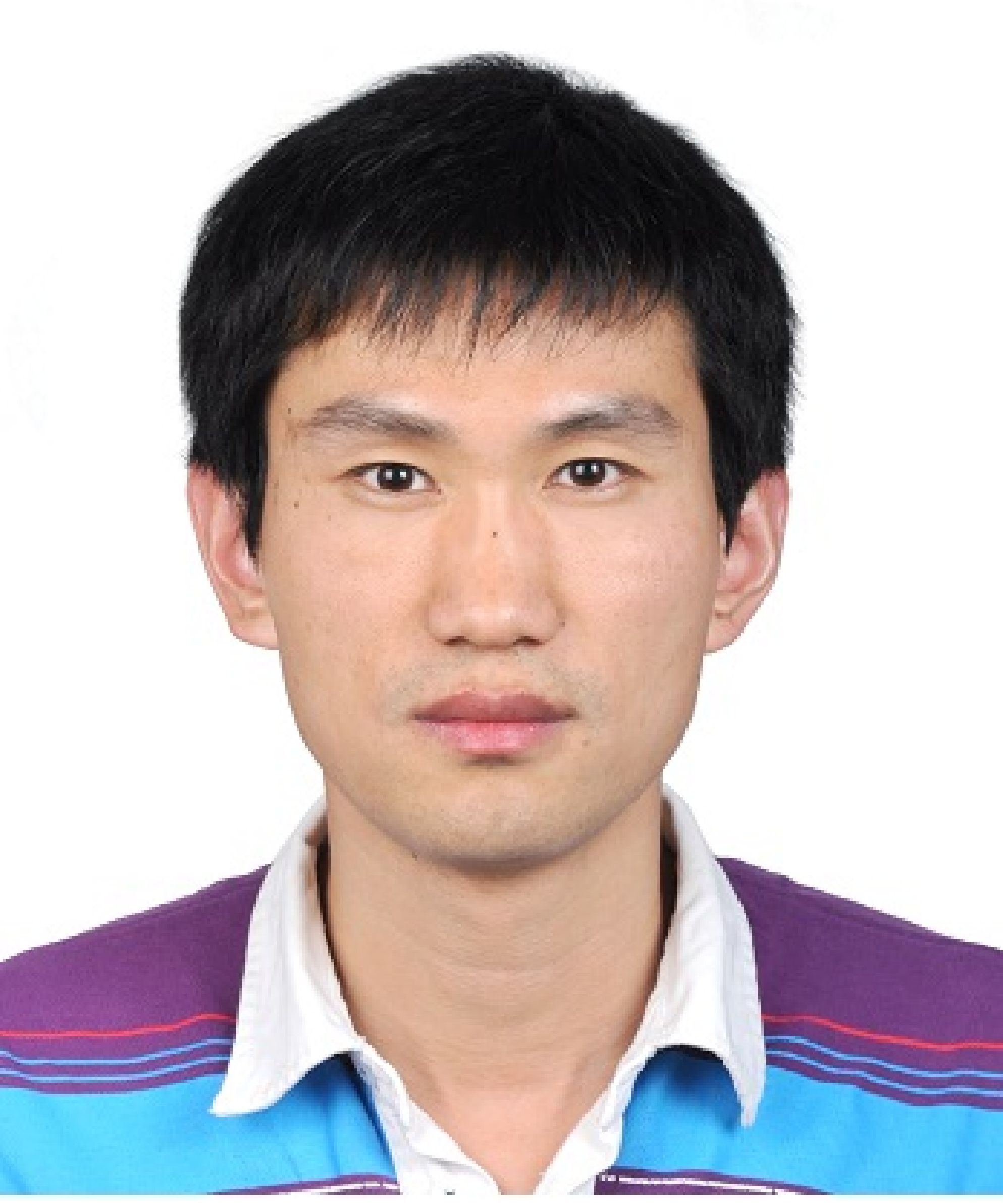}}]{Zhen Cui} received the B.S., M.S., and Ph.D. degrees from Shandong Normal University, Sun Yatsen University, and Institute of Computing Technology (ICT), Chinese Academy of Sciences in 2004, 2006, and 2014, respectively. He was a Research Fellow in the Department of Electrical and Computer Engineering at National University of Singapore (NUS) from 2014 to 2015. He also spent half a year as a Research Assistant on Nanyang Technological University (NTU) from Jun. 2012 to Dec. 2012. Currently, he is a Professor of Nanjing University of Science and Technology, China. His research interests mainly include deep learning, computer vision and pattern recognition.
\end{IEEEbiography}

\end{document}